%% file: arxiv.tex
\documentclass[10pt,twocolumn,letterpaper]{article}

\usepackage{cvpr}              

\input{preamble}
\usepackage{array}
\usepackage{multirow}
\usepackage{makecell} 
\usepackage{threeparttable}
\usepackage[ruled]{algorithm2e} 

\definecolor{cvprblue}{rgb}{0.21,0.49,0.74}
\usepackage[pagebackref,breaklinks,colorlinks,citecolor=cvprblue]{hyperref}

\def\paperID{126} 
\def\confName{3DV\xspace}
\def\confYear{2025\xspace}

\title{ShapeSplat: A Large-scale Dataset of Gaussian Splats and Their Self-Supervised Pretraining}

\author{  
\thanks{Indicates equal contribution. Author names are listed alphabetically.} Qi Ma\textsuperscript{1,2}\quad 
\footnotemark[1] Yue Li\textsuperscript{3}\quad 
\thanks{Corresponding author: Bin Ren, bin.ren@insait.ai} Bin Ren\textsuperscript{2,4,5}\quad
Nicu Sebe\textsuperscript{5}\quad 
Ender Konukoglu\textsuperscript{1}\quad \\
Theo Gevers\textsuperscript{3}\quad 
Luc Van Gool\textsuperscript{1,2}\quad
Danda Pani Paudel\textsuperscript{2}\\
\textsuperscript{1}Computer Vision Lab, ETH Zurich\quad 
\textsuperscript{2}INSAIT, Sofia University \\ \textsuperscript{3}University of Amsterdam\quad 
\textsuperscript{4}University of Pisa\quad 
\textsuperscript{5}University of Trento  
}

\begin{document}
\maketitle
\input{sections/0_abstract_arxiv}    
\input{sections/1_introduction}
\input{sections/2_relatedworks}
\input{sections/3_dataset}

\input{sections/4_method}
\input{sections/5_experiments}

\input{sections/6_conclusion}
\input{sections/7_supplmentary}

\clearpage
{
    \small
    \bibliographystyle{ieeenat_fullname}
    \bibliography{main}
}

\end{document}

%% file: preamble.tex
\usepackage[dvipsnames]{xcolor}
\usepackage{nomencl} 
\usepackage{tabularx}
\usepackage{xcolor}
\usepackage{etoolbox} 
\usepackage{titlesec}

\newcommand{\ours}{Gaussian-MAE\xspace}
\newcommand{\ourdata}{ShapeSplat\xspace}
\newcommand{\boldparagraph}[1]{\vspace{0.1em}\noindent{\bf #1}}

\def\eg{\emph{e.g.}\xspace} 
\def\ie{\emph{i.e.}\xspace} 

\def\wrt{\emph{w.r.t.}\xspace}
\def\etal{\emph{et al.}\xspace}

\titleformat{\section}[hang]
  {\normalfont\fontsize{11.96}{14.35}\bfseries} 
  {\thesection}                
  {1em}                       
  {}                          

\titleformat{\subsection}[hang]
  {\normalfont\fontsize{10.96}{13.15}\bfseries}
  {\thesubsection}
  {1em}{}

\titlespacing*{\section}{0pt}{1ex}{1ex}
\titlespacing*{\subsection}{0pt}{1ex}{0.5ex}

\usepackage[table,dvipsnames]{xcolor} 
\usepackage{colortbl} 
\colorlet{colorFst}{Green!25}       
\colorlet{colorSnd}{SpringGreen!45} 
\colorlet{colorTrd}{Yellow!30}      
\colorlet{colorLow}{darkgray!30}    
\definecolor{mygreen}{HTML}{D6C41C}
\definecolor{myred}{HTML}{FFB6C1}
\definecolor{myblue}{HTML}{9EB4F3}

\newcommand{\fs}{\cellcolor{colorFst}\bf}   
\newcommand{\nd}{\cellcolor{colorSnd}}      
\newcommand{\rd}{\cellcolor{colorTrd}}      

\newcolumntype{C}{>{\centering\arraybackslash}X}
\newcolumntype{R}{>{\raggedright\arraybackslash}X}

\makenomenclature
\renewcommand\nomgroup[1]{%
  \item[\bfseries
  \ifstrequal{#1}{G}{Gaussian parameters}{%
  \ifstrequal{#1}{S}{Scalars}{%
  \ifstrequal{#1}{M}{Method variables}{}}}%
]}

\newcommand{\nomenclcustom}[3]{\nomenclature[#1#2]{#3}}

\usepackage{tocloft}

%% file: sections/0_abstract_arxiv.tex
\begin{abstract}
3D Gaussian Splatting (3DGS) has become the de facto method of 3D representation in many vision tasks. This calls for the 3D understanding directly in this representation space. To facilitate the research in this direction, we first build a large-scale dataset of 3DGS using the commonly used ShapeNet and ModelNet datasets. Our dataset \textbf{\textit{ShapeSplat}} consists of \textbf{206K} objects spanning over 87 unique categories, whose labels are in accordance with the respective datasets. The creation of this dataset utilized the compute equivalent of 3.8 GPU years on a TITAN XP GPU.

We utilize our dataset for unsupervised pretraining and supervised finetuning for classification and segmentation tasks. To this end, we introduce \textbf{\textit{Gaussian-MAE}}, which highlights the unique benefits of representation learning from Gaussian parameters. Through exhaustive experiments, we provide several valuable insights. In particular, we show that  (1) the distribution of the optimized GS centroids significantly differs from the uniformly sampled point cloud (used for initialization) counterpart; (2) this change in distribution results in degradation in classification but improvement in segmentation tasks when using only the centroids; (3) to leverage additional Gaussian parameters, we propose Gaussian feature grouping in a normalized feature space, along with splats pooling layer, offering a tailored solution to effectively group and embed similar Gaussians, which leads to notable improvement in finetuning tasks. Our code will be available at \href{https://unique1i.github.io/ShapeSplat_webpage/}{ShapeSplat}. 
\end{abstract}

%% file: sections/1_introduction.tex
\section{Introduction}
\label{sec:introduction}

\begin{figure}[t]
\centering
        \centering
        \includegraphics[width=\linewidth]{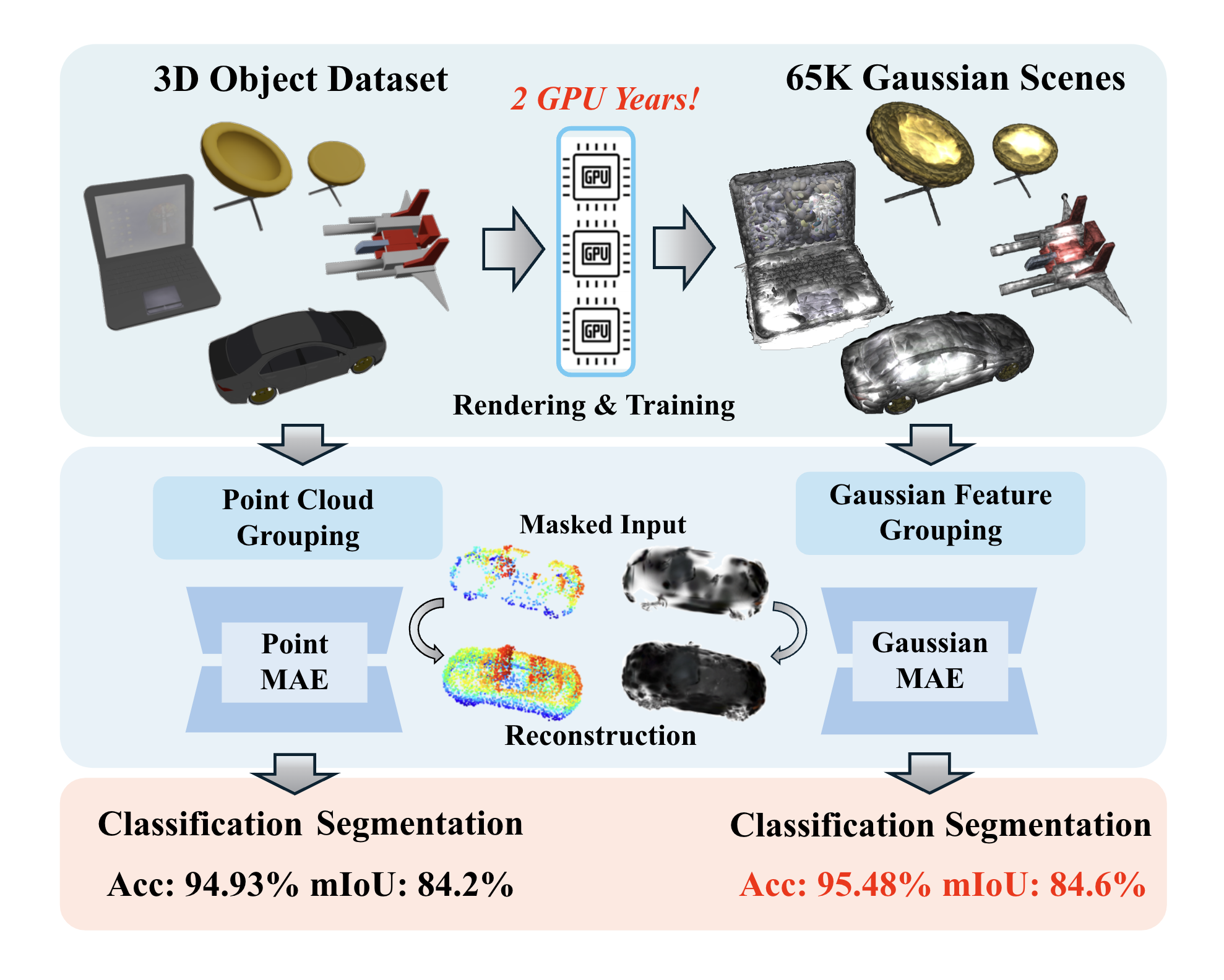}
\caption{\textbf{ShapeSplat Dataset and Gaussian-MAE.} We present \textit{ShapeSplat}, a large-scale dataset that took 3.8 GPU years to render and rasterize. Thanks to the dataset and the proposed pretraining method Gaussian-MAE, we outperform the point cloud counterpart method by 0.55$\%$ in accuracy and 0.4$\%$ in mean IoU.}
\label{fig:teaser}
\end{figure}

3D Gaussian Splatting~\cite{kerbl20233d}, a recent advancement in radiance field that represents 3D scenes using Gaussian primitives, has garnered significant research interest beyond view synthesis task, including scene reconstruction and editing~\cite{xie2024physgaussian, zhang2024rade}, segmentation and understanding~\cite{ye2023gaussian, qin2024langsplat}, digital human~\cite{qian2023gaussianavatars, lei2024gart}, and 3D generation~\cite{yi2024gaussiandreamer, tang2024lgm}.

Gaussian splats representation offers numerous advantages, including rapid rendering speeds, high fidelity, differentiability, and extensive editability. These features establish 3D Gaussian Splatting as a potential game changer for encoding the scenes in 3D vision pipelines.

To the best of our knowledge, there has been no exploration of direct learning on trained parameters of Gaussian splats in current research. A significant barrier in this regard is the lack of a large-scale dataset of trained Gaussian Splatting scenes. Despite 3DGS having considerably reduced the computation time, generating such a large-scale dataset remains significantly time-consuming. This paper seeks to address this challenge with the \textit{ShapeSplat} dataset, which comprises over 206K Gaussian splatted objects across 87 different categories. The dataset is designed to facilitate research in self-supervised pretraining of Gaussian splats, and to support downstream tasks, including classification, and 3D part segmentation, as shown in Fig.~\ref{fig:teaser}.

With this large-scale dataset, we first perform unsupervised pretraining in a masked autoencoder manner \cite{he2022masked}, followed by supervised finetuning for classification and segmentation tasks. Our extensive experiments demonstrate that each Gaussian parameter—opacity, scale, rotation, and spherical harmonics—can be effectively reconstructed during the pretraining stage. Moreover, incorporating additional Gaussian parameters notably enhances performance in downstream tasks.


We apply our model (pretrained on Gaussian centroids) to the uniformly sampled point clouds, showcasing its generalizability. 
We observed that while using Gaussian centroids alone is sufficient for the segmentation task, it leads to a performance drop in the classification task. This might be due to the complex spatial distribution of Gaussian parameters, as Gaussians tend to have larger scales and higher opacity in planar regions, while in corners and edges, they appear with smaller scale and opacity, as shown in \cref{fig:distribution_difference}. To address this, we propose a Gaussian feature grouping technique and splats pooling layer, whose effectiveness is demonstrated in pretraining and finetuning.

In summary, our key contributions are as follows: 

$\bullet$ We present \textit{ShapeSplat}, a large-scale Gaussian splats dataset spanning 206K objects in 87 unique categories. 

$\bullet$ We propose \textit{Gaussian-MAE}, the masked autoencoder-based self-supervised pretraining for Gaussian splats, and analyze the contribution of each Gaussian attribute during the pretraining and supervised finetuning stages.

$\bullet$ We propose novel Gaussian feature grouping with splats pooling layer during the embedding stage, which are customized to the Gaussian splats' parameters, enabling better reconstruction and higher performance respectively.


%% file: sections/2_relatedworks.tex
\section{Related Work}
\label{sec:related-work}
\subsection{3D Object Datasets}

Since the rapid advancement of 3D deep learning, despite the challenges of collecting, annotating, and storing 3D data, significant progress has been made in 3D datasets~\cite{downs2022google,fu20203dfuture3dfurnitureshape,morrison2020egad,park2018photoshape,collins2022abo, zhou2016thingi10k,shapenet2015,wu20153d,wu2023omniobject3d,deitke2023objaverse,deitke2024objaverse,uy2019revisiting} in recent years, thanks to the dedicated efforts of the community. ShapeNet~\cite{shapenet2015} has been a foundational platform for modeling, representing, and understanding 3D shapes in 3D deep learning. It offers over 3 million textured CAD models, with a subset called ShapeNet-Core containing 52K models, organized by mesh and texture quality. Similarly, ModelNet~\cite{wu20153d} provides 12K CAD models, while OmniObject3D \cite{wu2023omniobject3d} includes 6K objects with more category variation. To further scale up, efforts like Objaverse~\cite{deitke2023objaverse} and Objaverse-XL~\cite{deitke2024objaverse} have been introduced, containing 800K and 10.2 million models respectively and with rich text annotation, significantly accelerating research in large-scale 3D understanding. 
We selected ShapeNet~\cite{shapenet2015} as our choice for a pretrained Gaussian splats dataset due to its extensive size. For validation datasets in classification and 3D segmentation tasks, we chose ModelNet and ShapeNet-Part~\cite{yi2016scalable}. Additionally, we included ScanObjectNN~\cite{uy2019revisiting} for real-world applications. These choices were made because they offer similar categories and provide well-established benchmarks for  3D deep learning methods.
Related to our work,~\cite{irshad2024nerf,ma2024implicit} build NeRF~\cite{mildenhall2020nerfrepresentingscenesneural} representation on ScanNet and Omniobject3D. Instead, we use Gaussian splats as 3D representation.

\begin{figure}[t]
    \centering
    \begin{subfigure}[t]{0.23\textwidth}
        \includegraphics[width=\textwidth]{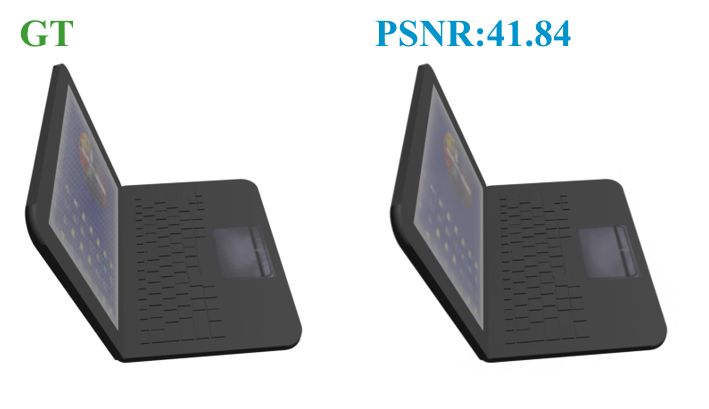}
        \caption{Laptop}
        \label{fig:Laptop}
        \vspace{3mm}
    \end{subfigure}
    \hfill
    \begin{subfigure}[t]{0.23\textwidth}
        \includegraphics[width=\textwidth]{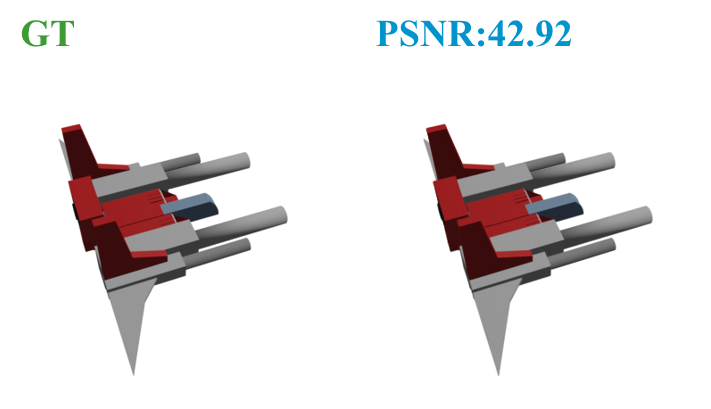}
        \caption{Airplane}
        \label{fig:Airplane}
    \end{subfigure}
    \hfill
    \begin{subfigure}[t]{0.23\textwidth}
        \includegraphics[width=\textwidth]{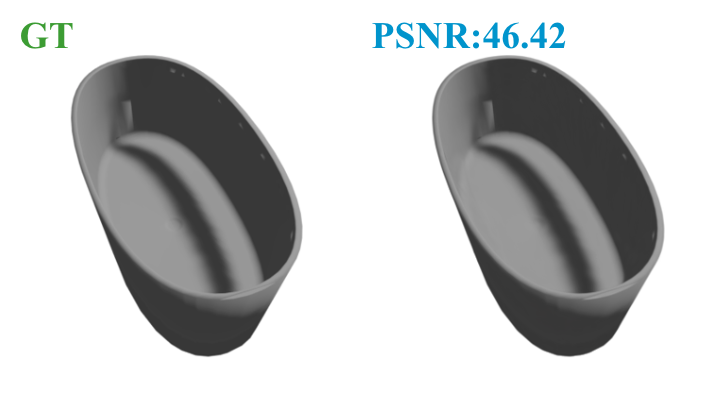}
        \caption{Bathtub}
        \label{fig:Bathtub}
    \end{subfigure}
    \hfill
    \begin{subfigure}[t]{0.23\textwidth}
        \includegraphics[width=\textwidth]{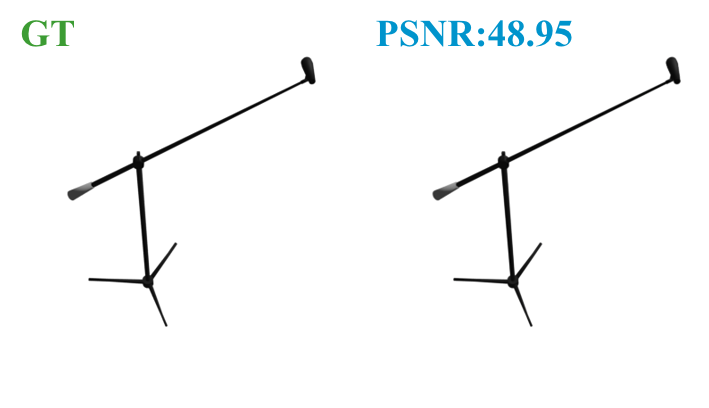}
        \caption{microphone}
        \label{fig:microphone}
    \end{subfigure}
    \hfill
    \vspace{1mm}
    \caption{\textbf{Qualitative Results of ShapeSplat Rendering}. We present high-quality renderings used for \ourdata on complex surfaces, shadows, and thin objects.}
    \label{fig:rendering}
\end{figure}

\input{tabs/data_statistics}

\subsection{Gaussian Splatting}
A plethora of research has focused on achieving better rendering quality, real-time speed, and scene editability with the advent of neural radiance fields, yet none have successfully addressed all these aspects. Building upon the rasterization technique, Kerbl \etal~\cite{kerbl20233d} proposed to represent the scene with a set of 3D Gaussian primitives, which significantly increased the rendering speed and obtained state-of-the-art rendering quality. Inspired by the success of 3D Gaussian Splatting (3DGS), a surge of following works further improve the rendering quality, geometry, and scene compression, extend to dynamic tasks, and utilize it as the representation bridging images and 3D scenes.

To represent accurate surface geometry,~\cite{guedon2024sugar,huang20242d,zhang2024rade} have introduced a rasterized method for computing depth and normal maps based on the Gaussian splats, and apply regularization for better surface alignment. Research on Gaussian scene compression~\cite{morgenstern2023compact, niedermayr2024compressed} utilized a learnable mask strategy or vector clustering technique to reduce total parameters without sacrificing rendering performance. Some works~\cite{luiten2024dynamic, yang2024deformable, lei2024mosca, gao2024gaussianflow} extended the 3D Gaussians to dynamic scenes and deformable fields by explicitly modeling the Gaussians across time steps or using a deformation network to decouple the motion and geometry structure. Attracted by the efficiency of 3DGS,~\cite{bortolon20246dgs, yugay2023gaussian, sandstrom2024splat} explored its optimization capability on re-rendering for pose estimation and mapping tasks. Researchers also have combined 3DGS with diffusion models to achieve efficient inpainting~\cite{liu2024infusion} and text-to-3D generation~\cite{tang2023dreamgaussian, yi2024gaussiandreamer}. Furthermore, there were efforts to lift 2D features from foundation models to attributes of Gaussian primitives~\cite{qin2024langsplat, zhou2024feature}, enabling language-driven probes and open-world scene understanding. Unlike these methods, our pipeline explores the rich information encoded in the Gaussian parameters without using any external features or supervision.

\subsection{Self-supervised 3D Representation Learning}
Self-supervised learning (SSL) has gained traction in computer vision due to its ability to generate supervisory signals from data itself~\cite{grill2020bootstrap,he2022masked,brown2020language}. In 3D representation learning, SSL methods using Vision Transformers (ViTs)~\cite{vaswani2017attention,dosovitskiy2020image,ren2023masked} are also advancing. These methods generally fall into two categories: contrastive and generative approaches. Contrastive pretraining focuses on learning discriminative representations to differentiate between samples, but applying this to ViTs-based 3D pretraining is challenging due to issues like overfitting and mode collapse~\cite{qi2023contrast,ren2024bringing}. In contrast, generative pretraining, inspired by the \textit{mask and reconstruct} strategy from BERT~\cite{devlin2018bert} and adapted to vision as masked autoencoder (MAE)~\cite{he2022masked}, has been extended to 3D by targeting point clouds~\cite{PointMAE,zhang2022point}, meshes~\cite{liang2022meshmae}, or voxels~\cite{hess2023masked}.

On this road, several works have advanced 3D representation learning by tailoring model architectures to specific input modalities, such as point clouds and voxel grids. Zhao \etal~\cite{zhao2021point} developed self-attention layers for point clouds that are invariant to permutations and cardinality. Yang \etal~\cite{yang2023swin3d} adapted the Swin Transformer~\cite{liu2021swin} for voxel grids, enabling scalability to large indoor scenes. Building on these architectures, some studies have explored encoding features from various scene representations using SSL. For instance, Yu \etal~\cite{zhang2022point} utilized a standard transformer and BERT-style pretraining with dVAE~\cite{rolfe2016discrete} for point patches. Irshad \etal~\cite{irshad2024nerfmae} applied MAE pretraining on Neural Radiance Fields, which enhanced scene-level tasks but faced challenges with scalability and data preparation. To the best of our knowledge, there is no prior work exploring the representation of trained 3D Gaussian splats parameters. 

%% file: tabs/data_statistics.tex
\begin{table}[t]
\centering
\resizebox{1\linewidth}{!}{%
\begin{tabular}{l|llll}
    \toprule
    Data Part & ShapeNet-GS & ShapeNetPart-GS & ModelNet-GS & Objaverse-GS \\
    \midrule 
    Task &   Classi. &   Part Seg.    &   Classi. &  Multi      \\
    Category &    55         &   50       &  40   & -      \\
    Objects &      52,121       &  16,823      &      12,308  &  141,703  \\
    GPU-days &    548.41         &  175.23       &     51.03  & 787.23    \\
    Gaussians &    24,267         &  23,689      &      22,456  & 50,000    \\
    PSNR &     44.19        &   44.54       &     45.10 & 33.71      \\
    JSD\cite{zyrianov2022learninggeneraterealisticlidar} &    0.231 &  0.207   &  0.230 & -    \\
    MMD\cite{zyrianov2022learninggeneraterealisticlidar} &   0.137 &  0.087   &  0.135 &  -    \\
    
    \bottomrule
    \end{tabular}}
    \caption{\textbf{Dataset Statistics}. We provide the specifications for the proposed ShapeSplat dataset, highlighting its quality and difference from the initial datasets in spatial distribution and the substantial preparation effort involved.
    }
    \label{tab:dataset_stat}
    \vspace{-2mm}
\end{table}

%% file: sections/3_dataset.tex

\begin{figure}[t]
\centering
        \centering
        \includegraphics[width=\linewidth]{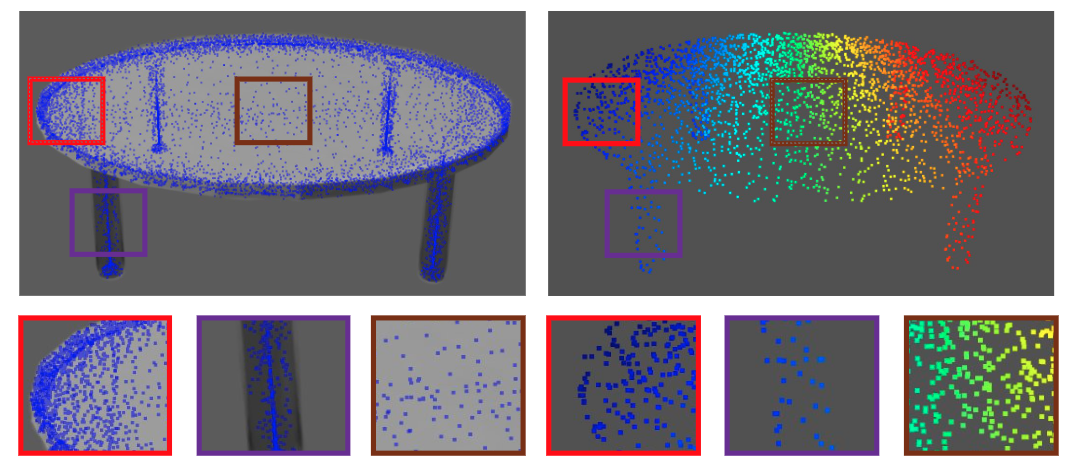}
\caption{\textbf{Distribution Comparison of Gaussian Splats Centroid and Point Cloud}. 
We highlight the difference between Gaussian centroids (left) and the point cloud used for initialization (right).}
\label{fig:distribution_difference}
\end{figure}

\begin{figure}[t]
    \centering
    \begin{subfigure}[t]{0.23\textwidth}
        \includegraphics[width=\textwidth]{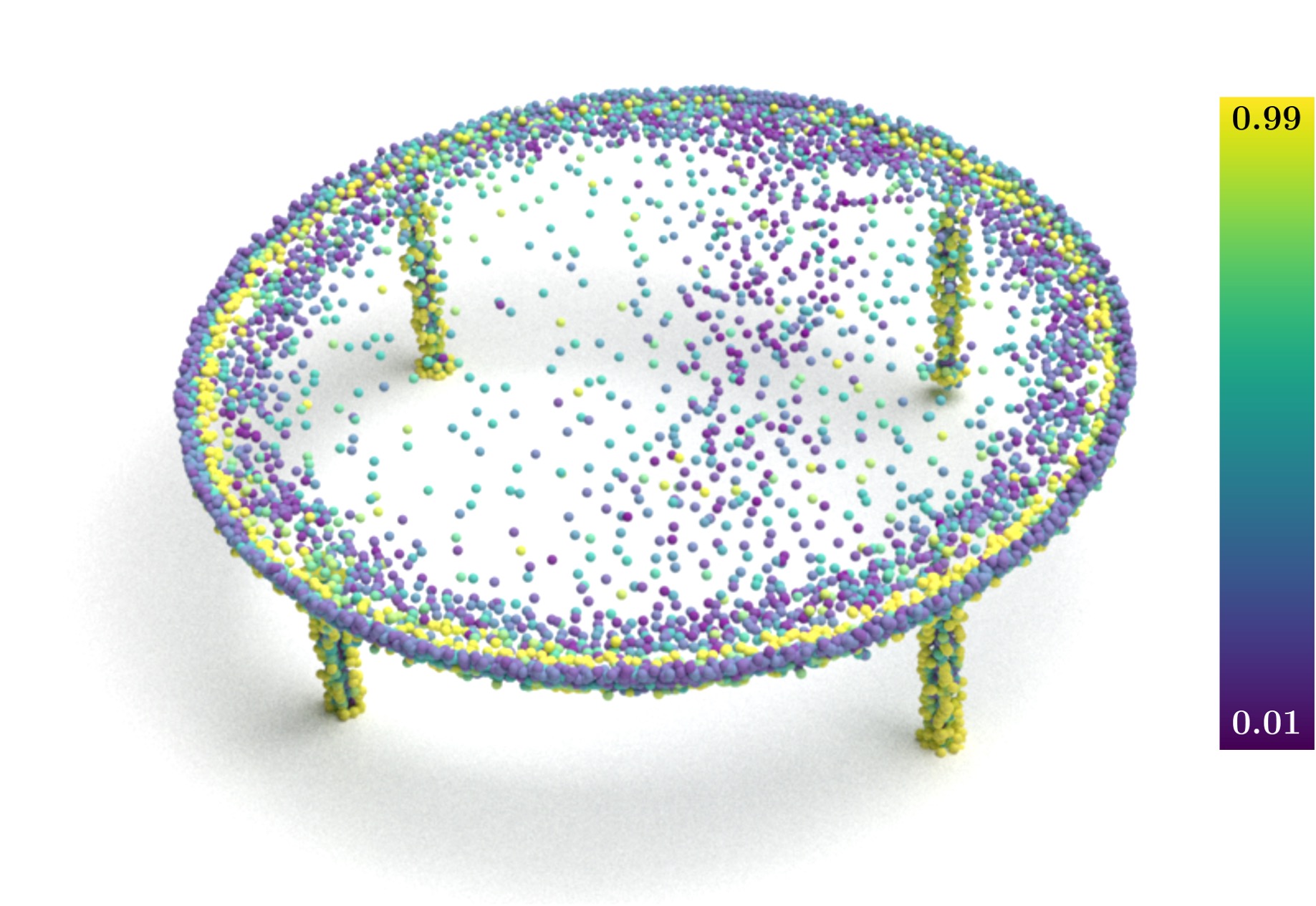}
        \caption{Opacity}
        \label{fig:vis_opacity}
    \end{subfigure}
    \hfill
    \begin{subfigure}[t]{0.23\textwidth}
        \includegraphics[width=\textwidth]{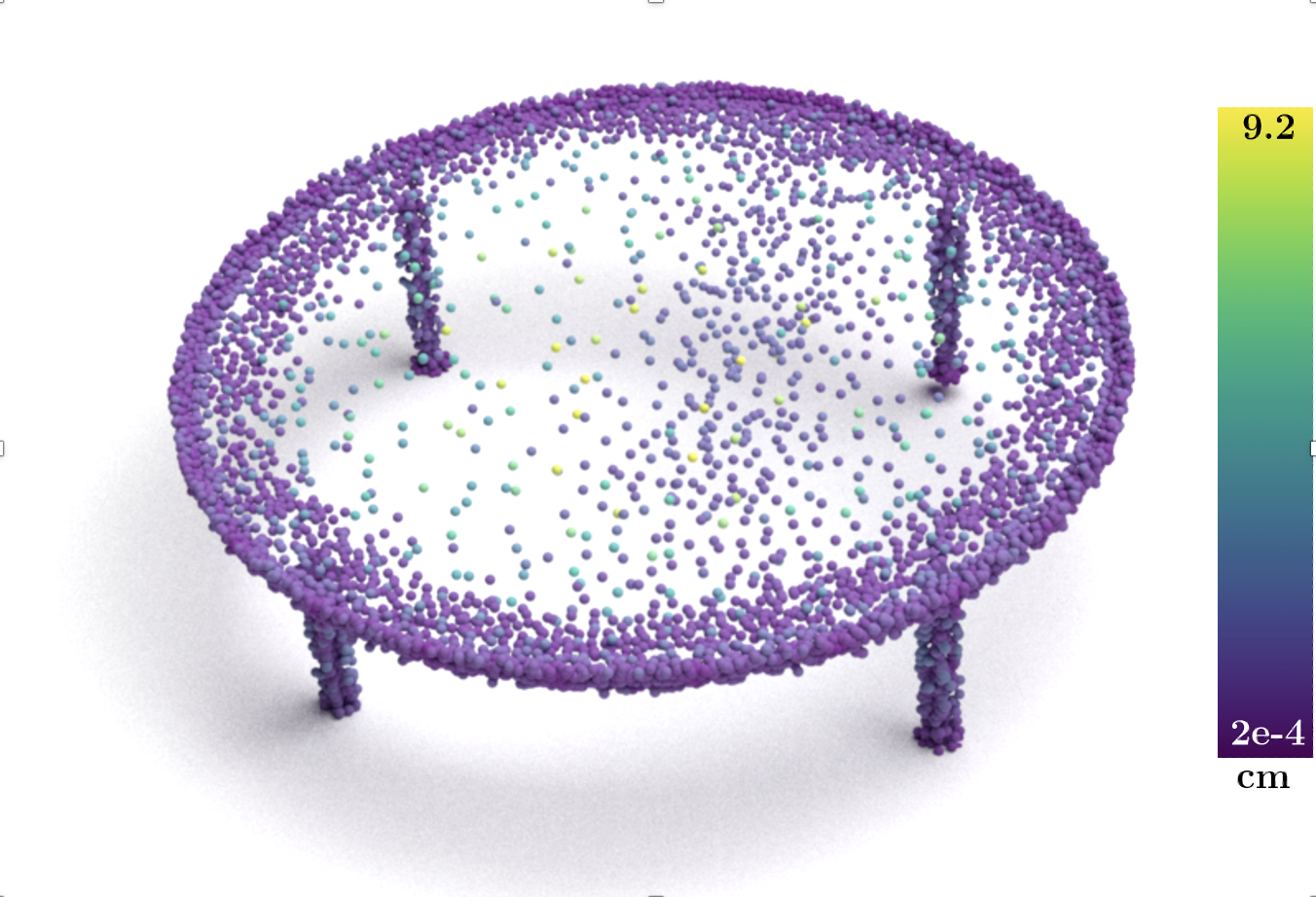}
        \caption{Scale}
        \label{fig:vis_scale}
    \end{subfigure}
    \hfill
    \begin{subfigure}[t]{0.23\textwidth}
        \includegraphics[width=\textwidth]{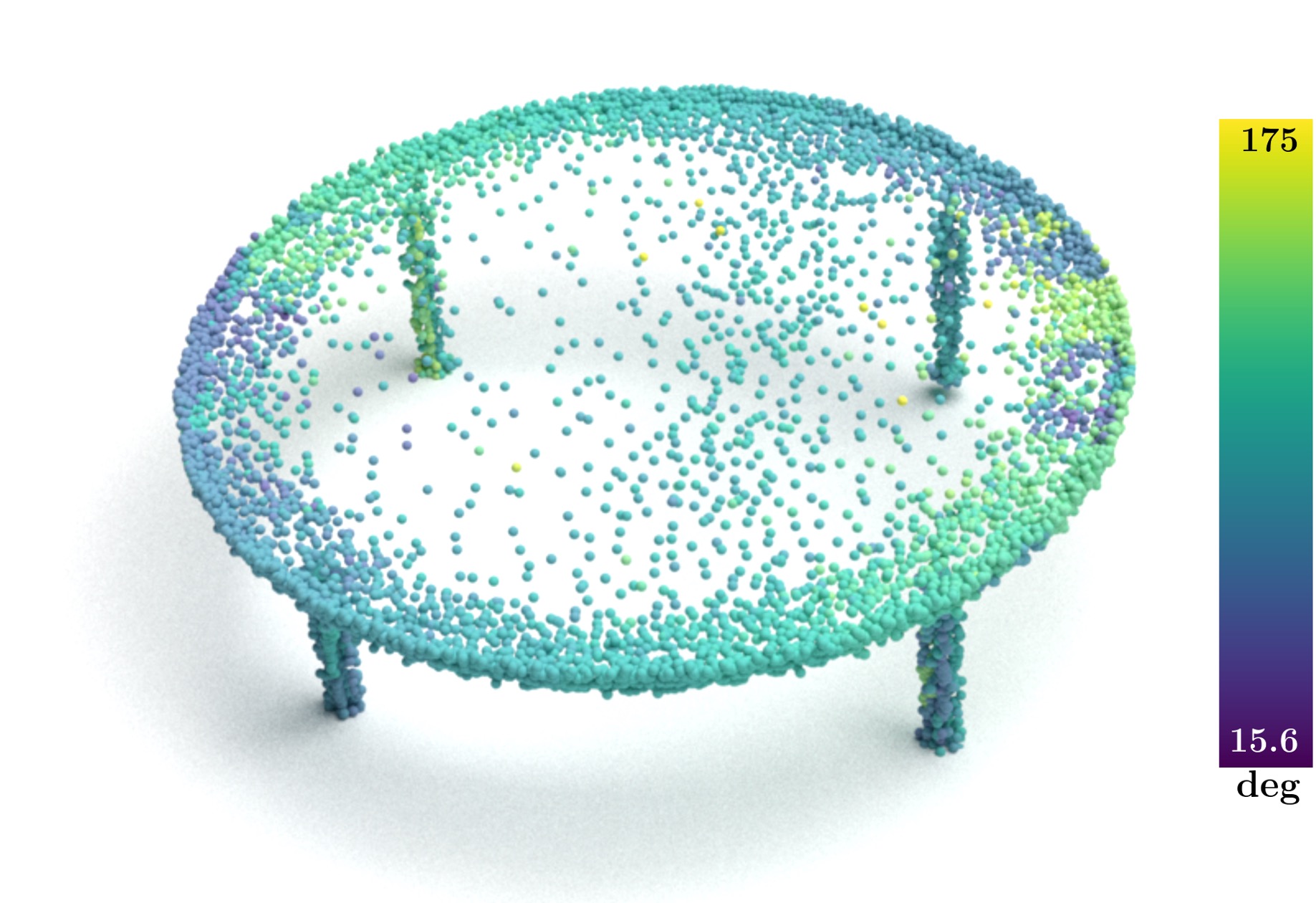}
        \caption{Rotation}
        \label{fig:vis_rotation}
    \end{subfigure}
    \hfill
    \begin{subfigure}[t]{0.23\textwidth}
        \includegraphics[width=\textwidth]{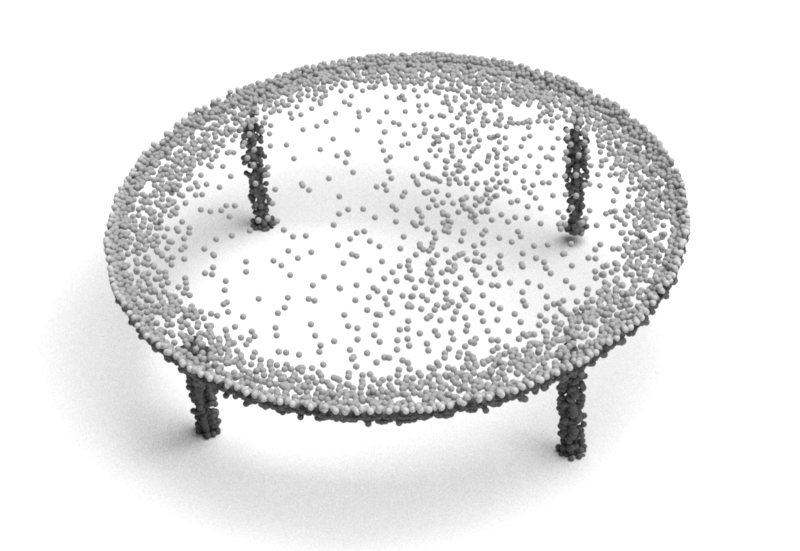}
        \caption{RGB from SH}
        \label{fig:vis_sh}
    \end{subfigure}
    \hfill
    \vspace{1mm}
    \caption{\textbf{Splats Centroid Colorized Based on Individual Gaussian Parameter Values}. The visualization reveals the complexity of the parameter distribution with respect to spatial dimensions.}
    \label{fig:attribute_vis}
    \vspace{-1mm}
\end{figure}

\begin{figure*}[t]
    \centering
    \includegraphics[width=1.0\linewidth]{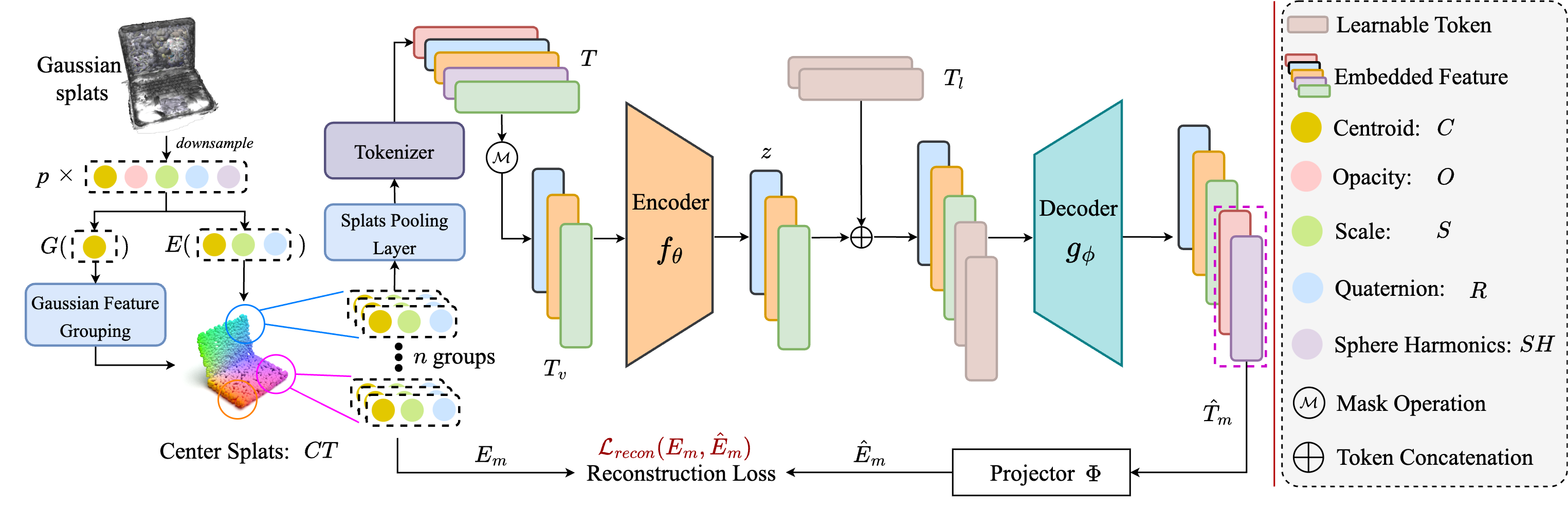}
    \caption{\textbf{\ours Framework}. Given the downsampled Gaussian splats, parameters for the grouping feature $G(\cdot)$ and embedding feature $E(\cdot)$ are first selected. $G(\cdot)$ is used to split the splats into $n$ groups, while $E(\cdot)$ serves as the reconstruction target in the MAE. After the splats pooling layer, the grouped splats are tokenized into group tokens $T$, which are then masked and the visible $T_v$ is passed into the encoder $f_{\theta}$ to obtain latent $z$. Following the concatenation of $z$ and the learnable tokens $T_l$, the decoder $g_{\theta}$ then recovers the masked tokens $\hat{T}_m$, which are projected to obtain the embedding components $\hat{E}_m$ for calculating the reconstruction loss $\mathcal{L}_\textit{recon}$. As an example, we use $G(C)$ for grouping and $E(C,S,R)$ for embedding, other inputs follow the same process.}
    \label{fig:pipeline}
\end{figure*}

\input{alg/algorithm}

\section{ShapeSplat Dataset Generation}
Since the datasets we choose are in CAD format, we first need to render 2D images at chosen poses to train Gaussian splats\cite{xu2019disn}. To address the limitations of the original 3DGS~\cite{kerbl20233d}, which often results in high redundancy~\cite{lu2024scaffold}, we incorporate important-score based pruning~\cite{fan2023lightgaussian}.

\boldparagraph{Model Rendering.} We adapt the code from~\cite{mildenhall2020nerfrepresentingscenesneural} to render CAD models in Blender. Each model is rendered from 72 uniformly spaced views, with an image resolution of 400$\times$400 to balance quality and rendering time. A white background is used for clear object distinction, and problematic CAD models are filtered out. \cref{fig:rendering} shows our qualitative rendering results.

\boldparagraph{3DGS Training.} After obtaining the image-pose pairs, we initialize the Gaussian centroids using a point cloud of 5K points uniformly sampled from mesh surfaces, following similar practice as in ShapeNet\cite{shapenet2015} and ModelNet\cite{wu20153d}. To reduce the scale artifacts, we employ the regularization term from \cite{yugay2023gaussian} to penalize large scales. The training process takes approximately 15 minutes per scene over 30,000 iterations. The total time required is detailed in \cref{tab:dataset_stat}.

\boldparagraph{Gaussian Pruning.} Following the approach in \cite{fan2023lightgaussian}, the importance score is calculated based on the contribution of each Gaussian splat to the rendered pixels, taking into account factors such as scale, rotation, and opacity. We prune 60 $\%$ of the Gaussians at iterations 16k and 24k. To avoid artifacts due to pruning, we incorporate pruning during the training process rather than as a post-processing step, ensuring necessary densification occurs throughout training. The final number of Gaussians is reported in \cref{tab:dataset_stat}.

\boldparagraph{Dataset Evaluation.} \cref{tab:dataset_stat} details the statistics of \ourdata, including object and category number, training time, number of Gaussians and the average PSNR (peak signal-to-noise ratio) of the renderings used for training. Note that ShapeNet-Part is a subset of ShapeNet-Core, and we report its training time for completeness. The average number of Gaussians exceeds 20K, notably higher than the input with 8K points in the point cloud baseline~\cite{PointMAE}.
Furthermore, we compute the Jensen–Shannon Divergence (JSD) and Maximum Mean Discrepancy (MMD)~\cite{zyrianov2022learninggeneraterealisticlidar} metrics between the trained Gaussian centroids and the initializing point cloud. These metrics are computed using 50x50 bins and averaged across three views, mapped onto the x, y, and z planes. Despite being initialized from the point cloud, the distribution of Gaussian splat centroids differs substantially. The visual differences in distribution are presented in~\cref{fig:distribution_difference}.

%% file: alg/algorithm.tex
\begin{algorithm}[htb]
\caption{Gaussian-MAE}
\label{alg:gaussian_mae}
\DontPrintSemicolon 
\KwIn{Object with $N$ Gaussian splats $X = [C, O, S, R, SH] \in \mathbb{R}^{N\times59}$}
\KwOut{Reconstructed embedding feature $\hat{E}_m$ for masked regions}
\BlankLine
  \begin{enumerate}
      \item Randomly downsample to $p$ Gaussian splats: $X' \in \mathbb{R}^{p\times59}$.
      \item Select parameters for the grouping feature $G(\cdot)$ and embedding feature $E(\cdot)$.
      \item Compute $n$ center splats from the grouping feature:
      \setlength{\abovedisplayskip}{4pt}
      \setlength{\belowdisplayskip}{4pt}
      $$
            CT = \operatorname{FPS}(G), \quad CT \in \mathbb{R}^{n\times f_{G}}.
      $$
      \item Find $k$ neighboring splats and obtain their embedding features:
      $$
            E_\text{group} = \operatorname{KNN}(E; G, CT), \quad E_\text{group} \in \mathbb{R}^{n\times k \times f_{E}}.
      $$
      \item Forward the embedding features to the tokenizer to obtain tokens:
      $$
            T = \operatorname{Tokenizer}(E_\text{group}), \quad T \in \mathbb{R}^{n \times D}.
      $$
      \item Masking with a ratio $r$, resulting in visible tokens $T_{v} \in \mathbb{R}^{(1-r)n \times D}$ , masked tokens $T_{m} \in \mathbb{R}^{rn \times D}$. and masked embedding feature $E_m$.
      \item Obtain the latent $z$ from the encoder: $z = f_{\theta}(T_{v})$.
      \item Concatenate $z$ with learnable token $T_l$ and then pass to the decoder $g_{\phi}(\cdot)$ to get the recovered masked token $\hat{T}_m=g_{\phi}(z \oplus T_l)$
      \item Projector $\Phi$ output the recovered embedding feature $\hat{E}_m = \Phi(\hat{T}_m)$ 
      \item Minimizing the reconstruction loss $\mathcal{L}_\text{recon}$:
      $$
        \min_{\theta, \phi} \underset{\substack{E \sim \mathcal{\tau}}}{\mathbb{E}} \left[\mathcal{L}_\text{recon} \left(E_m, \hat{E}_m\right)\right].
      $$
  \end{enumerate}
\end{algorithm}

%% file: sections/4_method.tex
\section{Gaussian-MAE}
\label{sec:method}
\noindent{\textbf{Preliminary: 3D Gaussian Splatting.}} 3DGS parameterizes the scene space with a set of Gaussian primitives $\{X_i\}_{i=1}^{N}$, stacking the parameters together, 
\begin{equation}
X = [C, O, S, R, SH] \in \mathbb{R}^{N\times59},
\end{equation}
with centroid $C \in \mathbb{R}^{N\times3}$, opacity $O \in \mathbb{R}^{N\times1}$, scale $S \in \mathbb{R}^{N\times3}$, quaternion vector $R\in \mathbb{R}^{N\times4}$ and sphere harmonics $SH \in \mathbb{R}^{N\times48}$. We refer to them as Gaussian features or Gaussian parameters. Each Gaussian can be seen as softly representing an area of 3D space with its opacity. A point $q$ in scene space is influenced by a Gaussian $X_i$ according to the Gaussian distribution value weighted by its opacity:
\begingroup
\begin{equation}
f_{i}(q) = O_i \exp\left( -\frac{1}{2} (q - C_{i})^T \Sigma_{i}^{-1} (q - C{i}) \right),
\end{equation}
\endgroup
where the covariance matrix $\Sigma_{i}$ can be formulated as $\Sigma_{i}=R_iS_iS_i^{T}R_i^{T}$. 

As the name "splatting" suggests, the influence function $f$ can be cast onto a 2D image plane for every Gaussian. The rendered color of a pixel is the weighted sum over the colors $c_i$ in its influencing Gaussians set $\mathcal{S}$, calculated by the alpha-blending equation using the sorted front-to-back order:
\begingroup
\setlength{\abovedisplayskip}{0pt}
\setlength{\belowdisplayskip}{0pt}
\begin{equation}
c_\text{pixel} = \sum_{i \in \mathcal{S}} c_i f^{\textrm{2D}}_{i} \prod_{j=1}^{i-1} (1 - f^{\textrm{2D}}_{j})\enspace.
\end{equation}
\endgroup
Through differentiable rasterization, the rendering loss is back-propagated to the trainable Gaussian parameters.

\subsection{Gaussian MAE Framework}
\cref{fig:pipeline} illustrates the pipeline of \ours. 
We report detailed algorithm in \cref{alg:gaussian_mae}. 
We define two key features: embedding feature $E$ ($E \subseteq X'$ with $E \in \mathbb{R}^{p \times f_{E}}$), a subset of $X'$ selected as input to the tokenizer, which is the reconstruction target, its dimension $f_{E}$ depends on the chosen parameters; 
and grouping feature $G$ ($G \subseteq X'$ with $G \in \mathbb{R}^{p \times f_{G}}$), another subset of $X'$ used to compute distances during the grouping process and its dimension $f_{G}$ depends on the chosen parameters. 
We use centroid grouping $G(C)$ for experiments in~\cref{tab:modelnet_cls,tab:domain,tab:partseg}. 
Multiple \texttt{Conv1d} projectors $\Phi$ are used for different gaussian parameters. Chamfer-Distance~\cite{fan2017point} is used for the reconstruction loss $\mathcal{L}_{recon}$ of centroid and $\mathcal{L}1$ is used for the rest.
\subsection{Gaussian Feature Grouping}
\label{subsec:gs_parameter}

Building on \cref{alg:gaussian_mae}, we propose considering more than centroid during grouping. The reasons for separating grouping and embedding are twofold: firstly, sampling only in centroid space assumes a similar correlation between all Gaussian parameters and its center coordinate, which is not the case. As shown in \cref{fig:attribute_vis}, Gaussian splats with varying opacity and scale can blend together on the surface. Secondly, using additional Gaussian features for grouping allows for the clustering of splats that are not only spatially close but also similar in other attributes.
To balance the grouping across gaussian parameters, we recenter all parameters (excluding quaternions) to zero mean and map them to the unit sphere in their respective dimensions. To avoid the complexity of grouping with high-dimensional features, we select only the $SH$ base $\in \mathbb{R}^{p\times 3} $ for grouping.

 

\subsection{Splats Pooling Layer}
Gaussian feature grouping effectively cluster Gaussians into groups by similarity. To further balance the contribution of each parameter, we propose a learned temperature-scaled splats pooling layer customed to Gaussian parameters, aiming to efficiently aggregate information from the embedded feature of the potential neighbors.


Let $B$ denote the batch size, $n$ the number of groups, $P$ the number of potential neighbors per group, $k$ the desired number of nearest neighbors, and $D$ the intermediate embedding dimension. For each center splat, $P$ neighbors are first selected by using KNN on centroids. Then, we apply \texttt{Conv1d} layers to obtain embedding $F$, where $F \in \mathbb{R}^{B \times n \times P \times D}$. For each group in the batch, the query item $y_{b,n}$ is obtained by applying max pooling to $F$ across the dimension $P$. We then compute the pairwise distances between the query item $y_{b,n}$ and potential neighbors as $d_{b,n,p} = \|y_{b,n} - F_{b,n,p}\|^2$ for $p \in [1, P]$.

To introduce learnable temperature, we define $\gamma \in \mathbb{R}^k$ and $\beta \in \mathbb{R}^k$, respectively. The temperatures are computed as $t = \exp(\gamma) + \beta$. Once the softmax weights are calculated by applying a temperature-scaled softmax to the pairwise distances, the resulting weight tensor is $W \in \mathbb{R}^{B \times n \times k \times P}$. These weights are used for feature aggregation, producing the aggregated tensor $Z \in \mathbb{R}^{B \times n \times k \times D}$, where $Z_{b,n,k,d} = \sum_{p=1}^{P} W_{b,n,k,p} \cdot F_{b,n,p,d}$. $Z$ is then served as the input to derive the token for the splats group.

Per-dimension temperature here can depend on the query item, enabling the model to learn when it is beneficial to average more uniformly across the embedding space with high temperature and with low temperature when it should focus on some distinct neighbor embeddings. 
This is especially needed as Gaussian attributes can exhibit highly uneven distribution as shown in \cref{fig:attribute_vis}. 

%% file: sections/5_experiments.tex
\section{Experiments}
\label{sec:experiments}
In this section, we extensively present the pretraining results, finetuning performance, generalization experiments, effect of Gaussian space grouping and splats pooling layer, as well as an ablation study. Please refer to our supplement for implementation details. 
\subsection{Pretraining Results}
\label{sec:pretrain}

We conduct extensive pretraining experiments with the proposed Gaussian-MAE method, demonstrating that Gaussian parameters can be successfully reconstructed. Detailed reconstruction analysis is reported in~\cref{fig:per_attribute_recon} and qualitative results are included in the supplement. In alignment with \cref{subsec:gs_parameter}, we use $E(\cdot)$ and $G(\cdot)$ for Gaussian parameters for embedding and grouping. To distinguish different models in self-supervised learning, we use \textbf{*} to indicate models pretrained on Gaussians and tested on point clouds, \textbf{$\dagger$} for models pretrained and finetuned on point clouds. Unless otherwise specified, we default to downsampling to 1024 Gaussians during pretraining and to 4096 during task-specific finetuning, and we use a mask ratio of 60\%.

\input{tabs/modelnet_cls}

\subsection{Classification Experiments}
To assess the efficacy of the pretrained model, we gauged its performance under three kinds of transfer protocols like~\cite{dong2022autoencoders,qi2023contrast,ren2024bringing} for classification tasks during finetuning, \ie, 
(a) \textbf{Full}: finetuning pretrained models by updating all backbone and classification heads. 
(b) \textbf{MLP-Linear}: The classification head is a single-layer linear MLP, and we only update these head parameters during finetuning.
(c) \textbf{MLP-3}: The classification head is a three-layer non-linear MLP (\ie, the same as the
one used in FULL), and we only update these head parameters during finetuning.

We report classification results on ModelNet10 and ModelNet40 datasets in \cref{tab:modelnet_cls}. 
Note that we use $G(C)$ for grouping and analyzing how different attributes contribute to this tasks. 
Compared to the supervised-only method, our \ours effectively learns shape priors from unlabeled data. Even when using only the Gaussian centroids, our results surpass those of the supervised-only methods. However, we observe that when using only the center positions of the Gaussians, our results are inferior to the unsupervised baseline, Point-MAE, on point clouds. This may be due to the uneven distribution of the Gaussians. Notably, when we use other Gaussian parameters, our performance improves significantly by $0.55\%$. Each component contributes positively to the results, with the scale and rotation components providing the greatest benefit to classification. Furthermore, the performance increase over the baseline becomes even larger when using linear ($0.66\%$) and MLP-3 ($1.65\%$) probing, suggesting that the pretrained features learned by our approach are stronger and more robust.

\subsection{Segmentation Experiments}
Following the same setting as point cloud pretraining methods like Point-Bert~\cite{PointBERT}, Point-MAE~\cite{PointMAE}, we use the pretrained weight from ShapeNet and finetine on ShapeNet-Part for the part segmentation task.
From the results reported in~\cref{tab:partseg}, we observed that employing only Gaussian centroid resulted in higher class mIoU compared to Point-MAE, indicating that pretraining on Gaussian centroids offers greater benefits for segmentation. This aligns with our previous findings, as Gaussian distributions, concentrated at boundaries, capture crucial semantic variations. Unlike classification, utilizing opacity, scale, and rotation improves segmentation results, while adding spherical harmonics (SH) negatively impacts performance.

\begin{figure*}[tbh!]
    \centering
    \centering
    \includegraphics[width=1.0\linewidth]{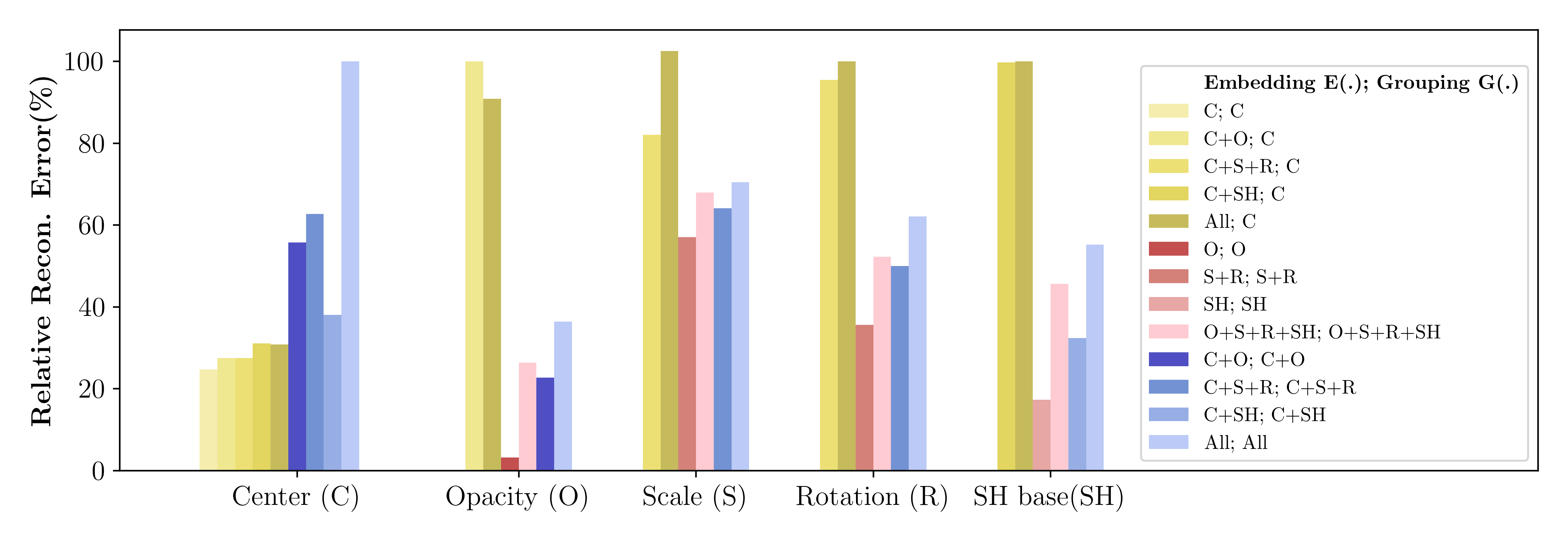}
    \vspace{-3mm}
    \caption{\textbf{Per-attribute Reconstruction of Different Grouping Method}. Relative reconstruction error is reported for each Gaussian parameter using different grouping feature $G$ represented by three different color shades, namely \textcolor{mygreen}{centroid grouping}, \textcolor{myred}{feature grouping without centroid}, and \textcolor{myblue}{feature grouping with centroid}. We observe that better overall reconstruction leads to better performance, \cf ~\cref{fig:ablation_gaussian_grouping}.}
    \label{fig:per_attribute_recon}
    \vspace{0mm}
\end{figure*}

\begin{figure}[htb]
\centering
        \centering
        \includegraphics[width=1.0\linewidth]{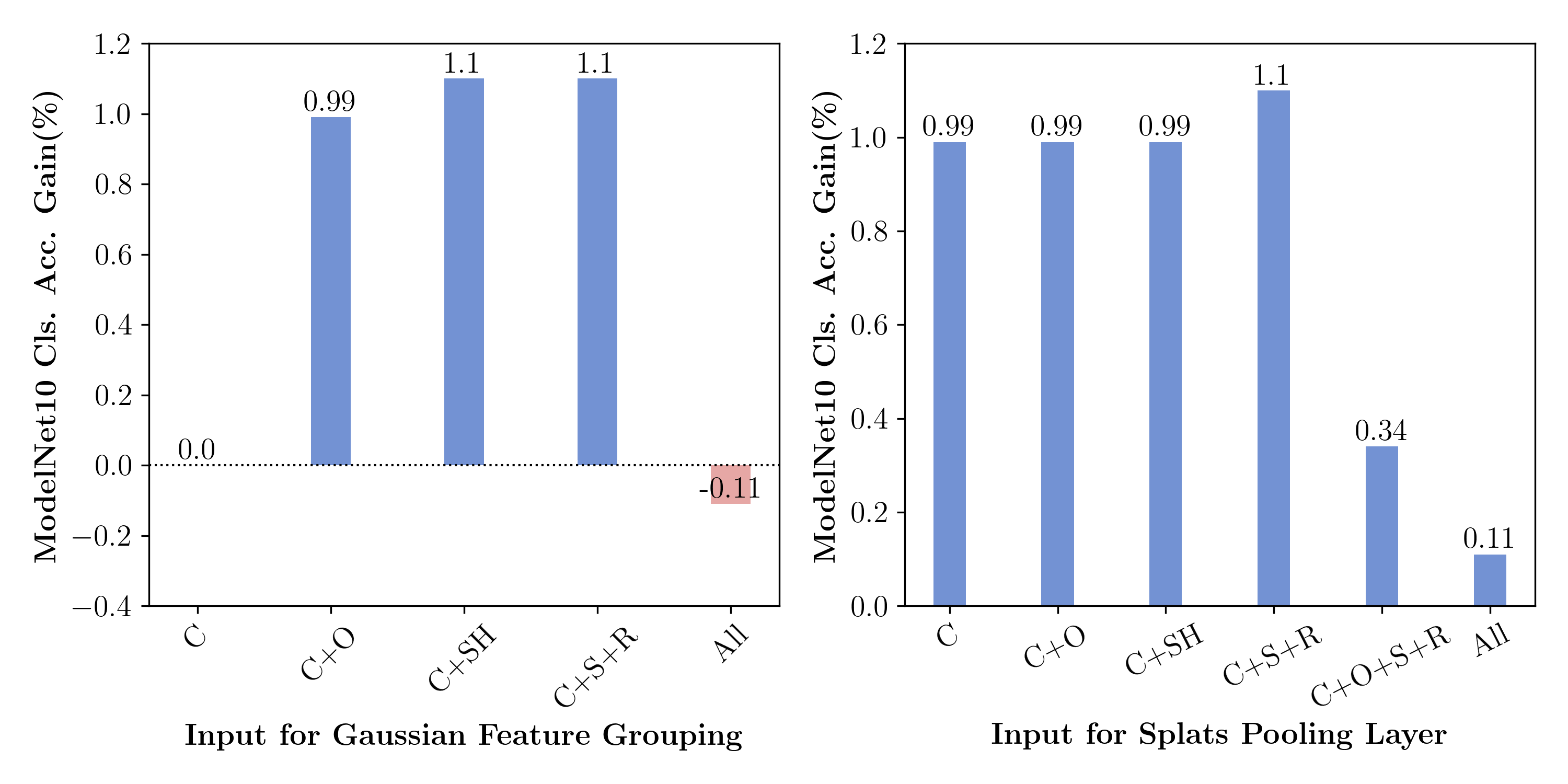}
\caption{\textbf{Ablation on Gaussian Feature Grouping and Splats Pooling Layer.} Classification accuracy on ModelNet10 significantly improves with the inclusion of both Gaussian feature grouping and splats pooling layer, across different embedding features.}
\label{fig:ablation_gaussian_grouping}
\end{figure}

\begin{figure}[htb]
\centering
        \centering
        \includegraphics[width=1.0\linewidth]{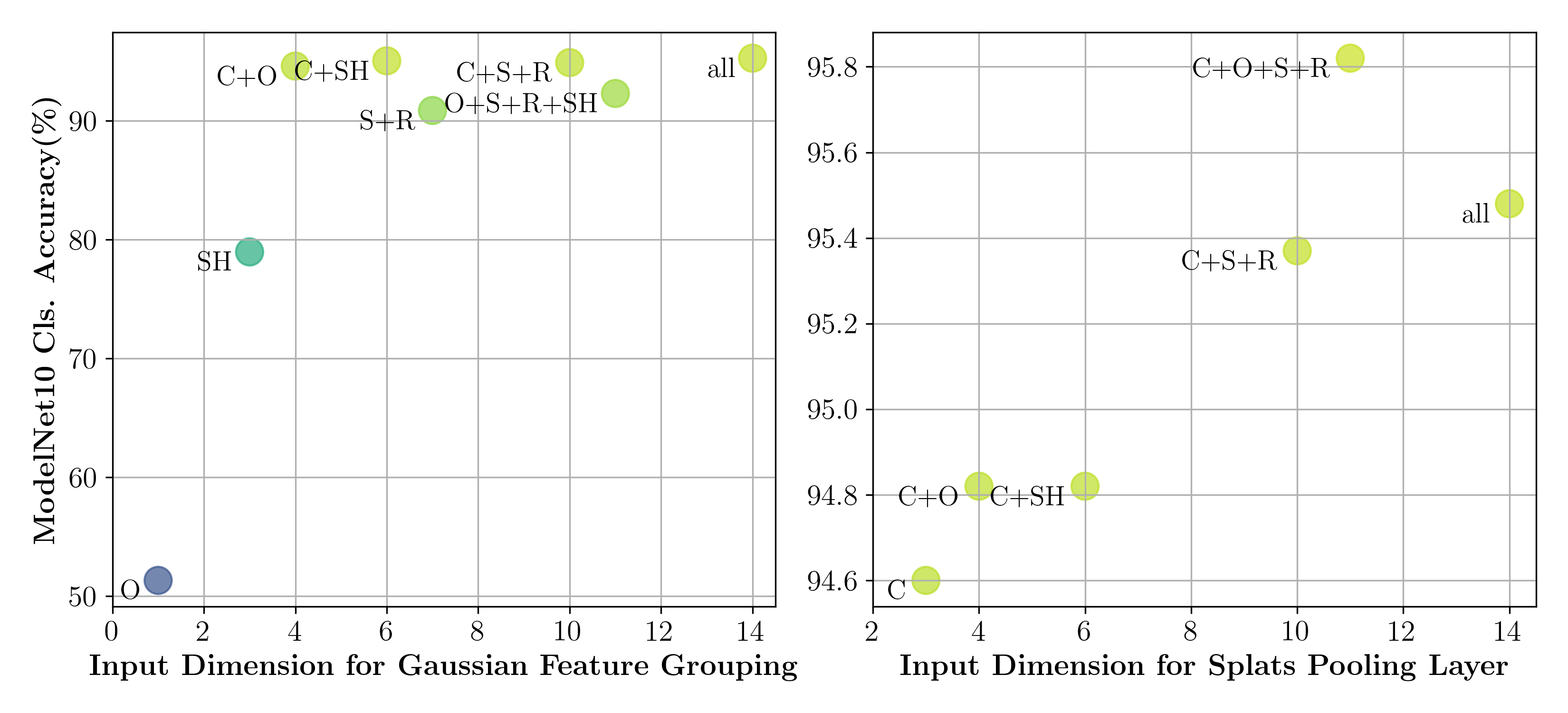}
\caption{\textbf{Effect of Input Dimension on Gaussian Feature Grouping and Splats Pooling Layer.} Higher input dimensions consistently improve classification accuracy with both modules.}
\label{fig:ablation_input_dimension}
\vspace{-2mm}
\end{figure}

\input{tabs/part_setmentation}
\subsection{Generalization to Point Clouds}

We report results of employing only the Gaussian centroids for pretraining followed by finetuning on point clouds in \cref{tab:domain}. Note that we tested on the point cloud data of ModelNet40 and ScanObjectNN. From the results, we can see that when joint training the encoder, the Gaussian-MAE pretraining on Gaussian centroids yields finetuning results close to Point-MAE pretraining on point clouds, and even outperforms the baseline in the Object-only results. However, when freezing the encoder and only training the decoder, the significant difference between the centroid distribution leads to a performance drop, and this gap becomes larger when using a Linear Classification head.

In the segmentation task reported in \cref{tab:partseg}, we were pleasantly surprised to find that the model pretrained on Gaussian centroids outperformed the point cloud baseline. This suggests that the encoder learned features from the Gaussian data that are more beneficial for segmentation. 


\subsection{Grouping Feature Grouping Analysis}
\label{sec:gs_space}
\boldparagraph{Effectiveness of Grouping Feature Grouping and Splats Pooling Layer.} We first evaluate the effect of grouping on multiple Gaussian parameters for ModelNet10 classification in \cref{fig:ablation_gaussian_grouping}. Compared to the baseline which groups only by Gaussian center, adding additional parameters individually leads to up to $1.1\%$ increase in classification accuracy. The same accuracy boost also results from the splats pooling layer, which uses the center for grouping but learns to combine multiple patterns for aggregating features from the group. For both methods, including all Gaussian parameters yields diminishing gains, as the baseline encodes stronger features from the parameter space. 

\boldparagraph{Reconstruction Analysis on Grouping Feature Grouping.} \cref{fig:per_attribute_recon} reports the relative reconstruction error of each Gaussian parameter with different grouping and embedding input. First, we observe that each feature achieves the lowest reconstruction error when it is both embedded and grouped exclusively. Additionally, grouping and embedding with other features have a negative impact on centroid reconstruction, while opacity, scale, and rotation affect more than SH as shown in the error bars for the center. This observation intuitively aligns with \cref{fig:attribute_vis}, as most colors in \textit{ShapeSplat} are uniform. We also observe that when grouped with centroid, each parameter has the highest reconstruction error, as indicated by the light yellow bars of opacity, scale, rotation, and SH. As shown by the rightmost light blue bars, when all the attributes are taken into account when grouping and embedded, the reconstruction errors are large and affect unevenly for different parameters, which we hypothesize results from the imbalance weighting of each feature. These observations highlight that Gaussian parameters are complexly distributed with respect to spatial dimensions. Furthermore, we conclude from \cref{fig:ablation_gaussian_grouping,fig:per_attribute_recon} that a smaller overall reconstruction error on the target embedding feature enhances performance in classification. 

\boldparagraph{Input Dimension Ablation for Parameter Space Grouping and Splats Pooling Layer.}
\cref{fig:ablation_input_dimension} ablates the total input dimension for Gaussian feature grouping and splats pooling layer with respect to the absolute classification accuracy on ModelNet10. We observe that for Gaussian feature grouping, more dimensions consistently correspond to higher accuracy, whereas for the splats pooling layer, it's generally the case. We obtain the highest accuracy of $95.82\%$ among our experiments using splats pooling layer with the input of center, opacity, scale, and rotation.

\subsection{Further Ablation Study}

\input{tabs/number_mask_ratio}

\boldparagraph{Ablation on Gaussian Splats Number for Finetuning.} \cref{tab:ablation_num_ratio} ablates the effect of the number of splats used during finetuning. For the classification task, the best performance occurs when the number of input splats matches that used in pretraining (1024 in this case), likely due to consistent aggregation patterns for patch embeddings. Accuracy improves again as the number of input splats increases further. For segmentation, the mIoU$_C$ generally improves with a higher number of input splats.

\boldparagraph{Ablation on Mask Ratio in Pretraining.} \cref{tab:ablation_num_ratio} reports the finetuning performance using the pretrained model with different mask ratios. The results show that as the mask ratio increases initially, both classification and segmentation performance improve, credited to the stronger representation learned by the MAE. As the mask ratio approaches 1, the pretraining target gets much more challenging, which in turn hinders the performance in downstream tasks.


%% file: tabs/modelnet_cls.tex
\begin{table}[t!]
    \centering
    \setlength\tabcolsep{7.5pt}
    \setlength{\extrarowheight}{0.8pt}
    \resizebox{\linewidth}{!}{
    \begin{threeparttable}
    \begin{tabular}{lcc}
    \toprule[0.95pt]
    Method & ModelNet10 & ModelNet40 \\
    \midrule[0.6pt]
    \multicolumn{3}{c}{\textit{Supervised Learning Only}}\\
    \midrule[0.6pt]
    PointNet~\cite{PointNet} & $\times$ & 89.2 \\
    PointNet++~\cite{PointNet++}  & $\times$ & 91.9 \\
    PTv1 \cite{zhao2021point} & $\times$ & 90.6 \\
    PTv2 \cite{wu2022pointtransformerv2grouped} & $\times$ & 91.6 \\
    \midrule[0.6pt]
    \multicolumn{3}{c}{\textit{with Standard ViTs Self-Supervised Learning} ({\scshape Full})}\\
    \midrule[0.6pt]
    Point-BERT$\dagger$~\cite{PointBERT} & 94.82 & \nd93.20 \\
    Point-MAE$\dagger$~\cite{pang2022maskedautoencoderspointcloud} & 94.93 & \nd93.20 \\
    Gaussian-MAE; $E(C)$ & 93.72  & 91.77  \\
    Gaussian-MAE;  $E(C,O)$& 93.83 & 91.78  \\
    Gaussian-MAE; $E(C,SH)$ & 93.83   & 92.41 \\
    Gaussian-MAE; $E(C,S,R)$& 94.27 &  93.19 \\
    Gaussian-MAE; $E(O,C,S,R)$ & \fs95.48 & 92.42  \\
    Gaussian-MAE; $E(All)$ & \nd95.37   & \fs93.35  \\

    \midrule[0.1pt]
    \multicolumn{3}{c}{\textit{with Standard ViTs Self-Supervised Learning} ({\scshape MLP-Linear})} \\
    \midrule[0.6pt]
    Point-BERT$\dagger$~\cite{PointBERT} & 93.06 &\nd 90.56 \\
    Point-MAE$\dagger$~\cite{pang2022maskedautoencoderspointcloud} &  93.17 & 90.24 \\

    Gaussian-MAE; $E(C)$ & 
    92.73  &  87.84  \\
    Gaussian-MAE; $E(C,O)$ & 91.30 & 87.43 \\
    Gaussian-MAE; $E(C,SH)$ &  91.19   &  86.38  \\
    Gaussian-MAE; $E(C,S,R)$ &  93.28 & 88.73  \\
    Gaussian-MAE; $E(O,C,S,R)$ & \nd93.72  &  88.93 \\
    Gaussian-MAE; $E(All)$  & \fs93.83   & \fs90.64  \\

    \midrule[0.6pt]
    \multicolumn{3}{c}{\textit{with Standard ViTs Self-Supervised Learning} ({\scshape MLP-3})} \\
    \midrule[0.6pt]
    Point-BERT$\dagger$~\cite{PointBERT} & \nd94.27 & \nd91.82 \\
    Point-MAE$\dagger$~\cite{pang2022maskedautoencoderspointcloud} & 93.61 & 92.63  \\
    Gaussian-MAE; $E(C)$ & 92.84  & 90.06  \\
    Gaussian-MAE; $E(C,O)$ & 93.39  & 89.86  \\
    Gaussian-MAE; $E(C,SH)$ & 93.39   & 90.72  \\
    Gaussian-MAE; $E(C,S,R)$ & 94.16   & 90.15  \\
    Gaussian-MAE; $E(O,C,S,R)$ & 93.72   & 91.29 \\
    Gaussian-MAE; $E(All)$ & \fs95.26   & \fs92.74  \\
    \bottomrule[0.95pt]
    \end{tabular}
    \end{threeparttable}
    }
    \caption{\textbf{Classification Accuracy on ModelNet} (overall accuracy $\uparrow$ [\%]). Best results are highlighted as \colorbox{colorFst}{\bf first}, \colorbox{colorSnd}{second}. $\dagger$ denotes the model pretrained and finetuned both on point clouds. Ours with all inputs as embedding feature $E$ yields the best accuracy.}
    \label{tab:modelnet_cls}
    \vspace{-2mm}
\end{table}

\begin{table}[ht]
    \centering
    \setlength\tabcolsep{7.5pt}
    \setlength{\extrarowheight}{1.5pt}
    \resizebox{\linewidth}{!}{
    \begin{threeparttable}
    \begin{tabular}{lcccc}
    \toprule[0.95pt]
    Method & ModelNet40 & OBJ\_BG & OBJ\_ONLY & PB\_T50\_RS\\
    \midrule[0.6pt]
    \multicolumn{5}{c}{\textit{Supervised Learning Only}}\\
    \midrule[0.6pt]
    PointNet~\cite{PointNet} & 89.2 & 73.3 & 79.2 & 68.0\\
    PointNet++~\cite{PointNet++}  & 91.9 & 82.3 & 84.3 & 77.9\\
    DGCNN~\cite{DGCNN} & 92.9 & 82.8 & 86.2 & 78.1 \\
    PointCNN~\cite{PointCNN} & 92.5 & 86.1 & 85.5 & 78.5\\

    \midrule[0.6pt]
    \multicolumn{5}{c}{\textit{with Standard ViTs Self-Supervised Learning} ({\scshape Full})}\\
    \midrule[0.6pt]
    Point-MAE$\dagger$~\cite{PointMAE} & \fs93.20 & \fs90.02 & 88.29 & \fs85.18 \\
    Gaussian-MAE*; $E(C)$ & 92.78  &  87.61 & \fs88.64 & 84.98\\

    \midrule[0.1pt]
    \multicolumn{5}{c}{\textit{with Standard ViTs Self-Supervised Learning} ({\scshape MLP-Linear})} \\
    \midrule[0.6pt]
    Point-MAE$\dagger$~\cite{PointMAE} &  \fs90.24 & \fs82.58 & \fs83.52  & \fs73.08 \\
    Gaussian-MAE*;$E(C)$ &  88.49 &  70.74  & 72.63 & 66.55 \\

    \midrule[0.6pt]
    \multicolumn{5}{c}{\textit{with Standard ViTs Self-Supervised Learning} ({\scshape MLP-3})} \\
    \midrule[0.6pt]
    Point-MAE$\dagger$~\cite{pang2022maskedautoencoderspointcloud} & \fs92.63 & \fs84.29  & 85.24  & \fs77.34  \\
    Gaussian-MAE*; $E(C)$ & 90.36 & 81.93  & \fs85.37  & 75.02  \\

    \bottomrule[0.95pt]
    \end{tabular}
    \end{threeparttable}
    }
    \caption{\textbf{Generalization evaluation on ModelNet and ScanObjectNN} (overall accuracy $\uparrow$ [\%]). * indicates models pretrained on Gaussians and tested on point clouds. Despite being pretrained on Gaussian splats, \ours achieves reasonable accuracy on point clouds, even surpassing the dedicated point cloud baseline in OBJ\_ONLY accuracy on ScanObjectNN. PB-T50-RS refers to the hard variant where ours did not lead.} 
    \label{tab:domain}
    \vspace{-3mm}
\end{table}

%% file: tabs/part_setmentation.tex
\begin{table}[t!]
    \centering
    \setlength\tabcolsep{3pt}
    \setlength{\extrarowheight}{0.3pt}
    \scalebox{0.80}{
    \begin{tabular}{lccc}
    \toprule[0.95pt]
    Method & mIoU$_C$ (\%) $\uparrow$ & mIoU$_I$ (\%) $\uparrow$  \\
    \midrule[0.6pt]
    \multicolumn{3}{c}{\textit{Supervised Representation Learning}}\\
    \midrule[0.6pt]
    PointNet~\cite{PointNet}  & 80.4 & 83.7   \\
    PointNet++~\cite{PointNet++} & 81.9 & 85.1  \\
    Transformer~\cite{vaswani2017attention} & 83.4 & 85.1  \\
    PTv1 \cite{zhao2021point} & 83.7 & 86.6  \\
    
    \midrule[0.6pt]
    \multicolumn{3}{c}{\textit{with Self-Supervised Representation Learning} }\\
    \midrule[0.6pt]
    Point-BERT$\dagger$~\cite{PointBERT} & 84.1 & 85.6 \\
    Point-MAE$\dagger$~\cite{pang2022maskedautoencoderspointcloud} & {84.2} & \fs86.1 &   \\
    Gaussian-MAE*; $E(C)$  & \fs84.6 & \fs86.1   \\

    Gaussian-MAE; $E(C)$ & 84.4  & 85.8   \\
    Gaussian-MAE; $E(C,O)$ &  \fs84.6 & \fs86.1   \\
    Gaussian-MAE; $E(C,SH)$  &  84.0 & 85.8  \\
    Gaussian-MAE; $E(C,S,R)$  & 84.4 & 86.0  \\
    Gaussian-MAE; $E(C,O,S,R)$  & 84.2  & 85.8  \\
    Gaussian-MAE; $E(C,All)$  & 84.4  & 86.0  \\

    \bottomrule[0.95pt]
    \end{tabular}
    }
    \caption{\textbf{Part Segmentation on \textbf{ShapeNet-Part}.} * indicates models pretrained on Gaussians and tested on point clouds. The class mIoU (mIoU$_C$) and the instance mIoU (mIoU$_I$) are reported. \ours yields $0.4\%$ gain in mIoU$_C$. }
    \label{tab:partseg}
\vspace{-5mm}
\end{table}

%% file: tabs/number_mask_ratio.tex

%

\begin{table}[t!]
    \centering
    \setlength\tabcolsep{7.5pt}
    \setlength{\extrarowheight}{1.5pt}
    \label{tab:ablation}
    \resizebox{\linewidth}{!}{
    \begin{threeparttable}
    \begin{tabular}{lccc}
    \toprule[0.95pt]
    Method & ModelNet10-GS & ModelNet40-GS & ShapeNetPart-GS \\
    \midrule[0.6pt]
    \multicolumn{4}{c}{\textit{splats input number ablation} ({\scshape Full})}\\
    \midrule[0.6pt]

    1024  & \fs95.37  & \fs93.35  & 84.4  \\
    2048 & 93.29  &  92.29   &  84.4 \\
    4096 & 94.82  &  93.02   &  \fs84.8 \\
    8192 & \nd95.26  & \nd93.05 & \nd84.5 \\

    \midrule[0.1pt]
    \multicolumn{4}{c}{\textit{mask ratio ablation} ({\scshape Full})} \\
    \midrule[0.6pt]
    mask ratio=0.2  & 95.48  & 91.73  & \nd84.6 \\
    mask ratio=0.4  & \fs96.03  & 91.89  &  \fs84.7 \\
    mask ratio=0.6  & 95.37  & \fs93.35  &  84.4  \\
    mask ratio=0.8  &  \nd95.81 & \nd92.46  & 83.0  \\

    \bottomrule[0.95pt]
    \end{tabular}
    \end{threeparttable}
    }
    \caption{\textbf{Ablation on Number of Splats and Mask Ratio.}
    }
\vspace{-2mm}
\label{tab:ablation_num_ratio}
\end{table}

%% file: sections/6_conclusion.tex
\section{Limitation}
\label{sec:limitation}
Compared to the original number of Gaussian Splats shown in \cref{tab:dataset_stat}, we significantly downsample them before pretraining, which mimics the processing of point clouds. 
This approach is sub-optimal because downsampling splats lose crucial details in appearance and geometry, impairing the representation. Exploring efficient methods to work directly with original splats could be a promising future direction.

\section{Conclusion}
\label{sec:conclusion}
We present \textit{ShapeSplat} dataset which enables the masked pretraining directly on 3DGS parameters. Experiments show that naively treating Gaussian centroids as point clouds does not perform well in downstream tasks. In contrast, our \textit{\ours} method excels by effectively aggregating features using Gaussian feature grouping and splats pooling layer. We hope this work opens a new avenue for self-supervised 3D representation learning.


%% file: sections/7_supplmentary.tex
\clearpage

\setcounter{page}{1}
\setcounter{section}{0}
\counterwithin{figure}{section} 
\counterwithin{table}{section}
\renewcommand{\thesection}{\Alph{section}} 
\maketitleappendix

\begin{abstract}
    This supplementary material first provides the nomenclature of our method for reference, followed by further information for better reproducibility as well as additional evaluations and qualitative results. 
\end{abstract}

\section{Nomenclature}
\vspace{-2em}
\input{tabs/nomenclature}
\printnomenclature

\section{Implementation Details}
\label{sec:implmentation}

\begin{figure}[ht!]
\centering
        \centering
        \includegraphics[width=0.9\columnwidth]{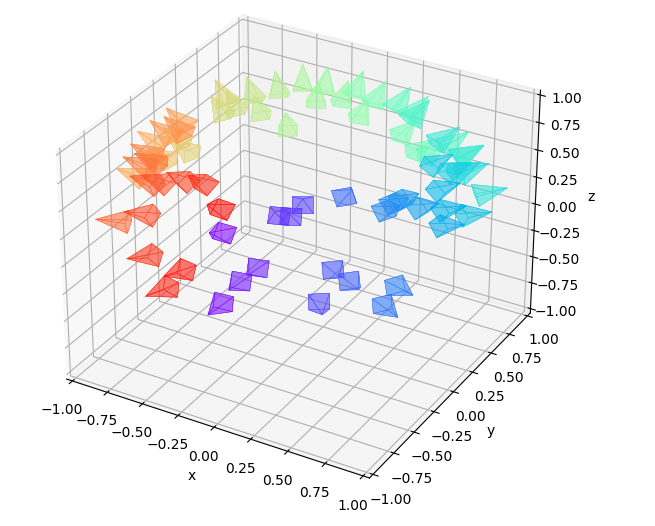}
\caption{\textbf{Sampled Views for \textit{ShapeSplat} rendering.}
We evenly select 72 views on the upper atmosphere for every object.}
\label{fig:model_rendering}
\end{figure}

\boldparagraph{Problematic CAD Model}: We excluded two models due to error 'No vertex found' encountered when loading into Blender: \texttt{4a32519f44dc84aabafe26e2eb69ebf4} and \texttt{67ada28ebc79cc75a056f196c127ed77}.
\boldparagraph{Model Rendering.} \cref{fig:model_rendering} illustrates the evenly chosen 72 views across the upper atmosphere when rendering a CAD model in ShapeNet~\cite{shapenet2015} and ModelNet~\cite{wu20153d}.

\boldparagraph{3DGS Initialization.}
To balance quality and speed, $5K$ points are sampled for 3DGS initialization. We employ a surface sampling method similar to the point cloud baseline. The normals of the sampled points are determined using the normals of their corresponding faces, and colors are assigned through interpolation from neighboring points, when available. For objects without material data, we initialize the color to gray.

\boldparagraph{Gaussian-MAE.} Gaussian-MAE model is configured with a masked encoder and a decoder, in which the masked encoder consists of the tokenizer and transformer encoder. The transformer encoder has a dimension of 384, a depth of 12 layers, and employs 6 attention heads per layer. The encoder uses a mask ratio of 0.6 and includes positional embeddings. The decoder shares the same dimension of 384 and utilizes 6 attention heads but with a shallower depth of 4 layers. Both the encoder and decoder are regularized using a drop path rate of 0.1. In addition, for the masked autoencoder (MAE), we use the $\ell_2$ Chamfer-Distance~\cite{fan2017point} following~\cite{PointMAE}. Let $\hat{E}_m = \Phi(\hat{T}_m)$ and $E_m$ be the reconstructed embedding feature and ground truth embedding feature, respectively. The reconstruction loss $ \mathcal{L}_{\text{recon}}$ can be written as:
      

\begin{equation}\label{eq:mpm}
    \mathcal{L}_{\text{recon}} =
    \sum_{} 
    \left[
        \begin{aligned}
            &\frac{1}{|{\hat{E}(C)}|} \sum_{{re}\in\hat{E}(C)} 
            \mathop{\min}\limits_{gt\in{E(C)}} \|re-gt\|^2_2  + \\
            &\sum_{gt\in{E(C)}}
            \mathop{\min}\limits_{re\in\hat{E}(C)} \|re-gt\|^2_2 + \\
            & \sum_{{re\in\hat{E}(O,S,R,SH)}} \|re-gt\|_1 
        \end{aligned}
    \right].
\end{equation}

\boldparagraph{Tokenizer.} The tokenizer for Gaussian parameters first utilizes an \texttt{Conv1d} layer that project the raw Gaussian parameters (embedding feature such as center, opacity, scale, rotation, and sh) into a 128-dimensional tensor. It then employ our splats pooling layer that effectively aggregate information from a larger neighborhood down to a smaller dimension. Finally, a second \texttt{Conv1d} followed by max pooing processes the features, resulting in a group token of dimension 384, which is then passed to the transformer encoder.

\boldparagraph{Parameter Numbers.} \cref{tab:model_params} summarizes the total and trainable parameter numbers for our pretraining and finetuning model, as well as the giga floating point operations (GFLOPs) and giga multiply-accumulate operations (GMACs). The pretraining model has the most trainable parameters of $28.79M$, while the linear probing in finetuning is the lightest with only $0.008M$ trainable parameters.

\boldparagraph{Training Details.} Experiments on pretraining were conducted using A6000 GPUs with a batch size of 128. For finetuning with 2048, 4096, and 8192 sampled splats, we used a batch size of 224 on H100 GPU. Both the pretraining and finetuning stages were trained over 300 epochs, utilizing a cosine learning rate scheduler with a 10-epoch warm-up period. The AdamW optimizer was employed, with learning rates set to 1 $\times 10^{-3}$ for pretraining and 5 $\times 10^{-4}$ for finetuning, alongside a weight decay of 0.05. More details are provided in Tab.~\ref{tab:hyper_params}.

\boldparagraph{Evaluation.} For the classification task, \textit{ShapeSplat} uses the labels from the original dataset. In the part segmentation task, to ensure a fair comparison with point cloud-based methods, our model predicts labels for the given point cloud positions. Features from all Gaussian centers are forwarded to the ground truth point cloud locations through distance-based interpolation, followed by feature integration using a \texttt{Conv1d} layer. We report the best results from the separate runs using checkpoints saved at epochs 250, 275, and 300 during pretraining.

As discussed in~\cite{van2023point}, the ModelNet40 dataset contains notable duplication and label errors that hinder model performance. The evaluation on ModelNet10, which is less affected by these issues, provides a more reliable benchmark and we assign more weight on it when drawing conclusions.

\input{tabs/model_params}
\input{tabs/hyper_paramters}

\section{Per-attribute Reconstruction Error}
\label{sec:3dgs_recon_error}
\input{tabs/gaussian_recon_error}
\cref{tab:recon_error} provides the detailed per-attribute reconstruction error corresponding to Figure 6 in the main paper. Note that the target reconstruction is determined by the choice of embedding feature, which results in blank areas in the table. Bundling other Gaussian attributes with the center into either the embedding feature or the grouping feature leads to improved overall reconstruction, and as evident from the classification results, better reconstruction then yields performance increase. We also present the results for \textit{Grouping in Gaussian Feature Space without Center}, where Gaussian centers are entirely excluded during pretraining and finetuning. This experiment shows that reasonable performance can still be achieved by relying solely on other parameters.

\section{Gaussian Splats Reconstruction}

\begin{figure}[ht]
\centering
\begin{tabularx}{\columnwidth}{XXX}
    \multicolumn{3}{c}{%
        \includegraphics[width=\linewidth]{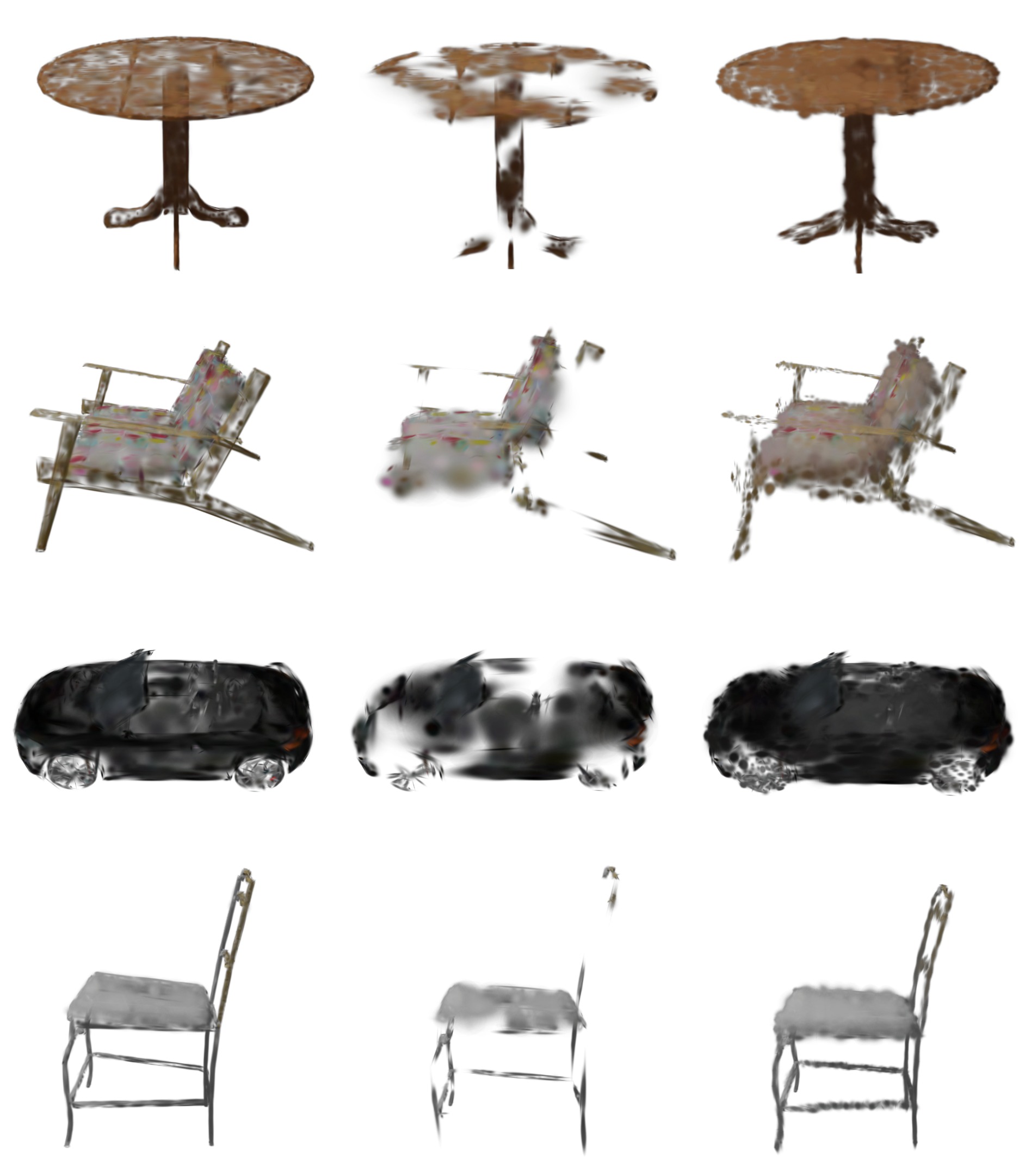}%
    } \\
    \hspace{0.4cm} \small (a) Sampled &  \small \hspace{0.4cm} (b) Masked & \small (c) Reconstructed \\
\end{tabularx} 

\caption{\textbf{Visualizations of Masked Inputs and Reconstructed Gaussians}. We show:
(a) Renderings of downsampled 4096 Gaussians, (b) Renderings with 60$\%$ Gaussians masked, (c) Renderings of reconstructed Gaussians. The pretrained \ours is able to reconstruct fine details, \eg, frames of the chair.}
\label{fig:vis_recon}
\end{figure}

\begin{figure*}[ht!]
\centering
        \centering
        \includegraphics[width=0.9\linewidth]{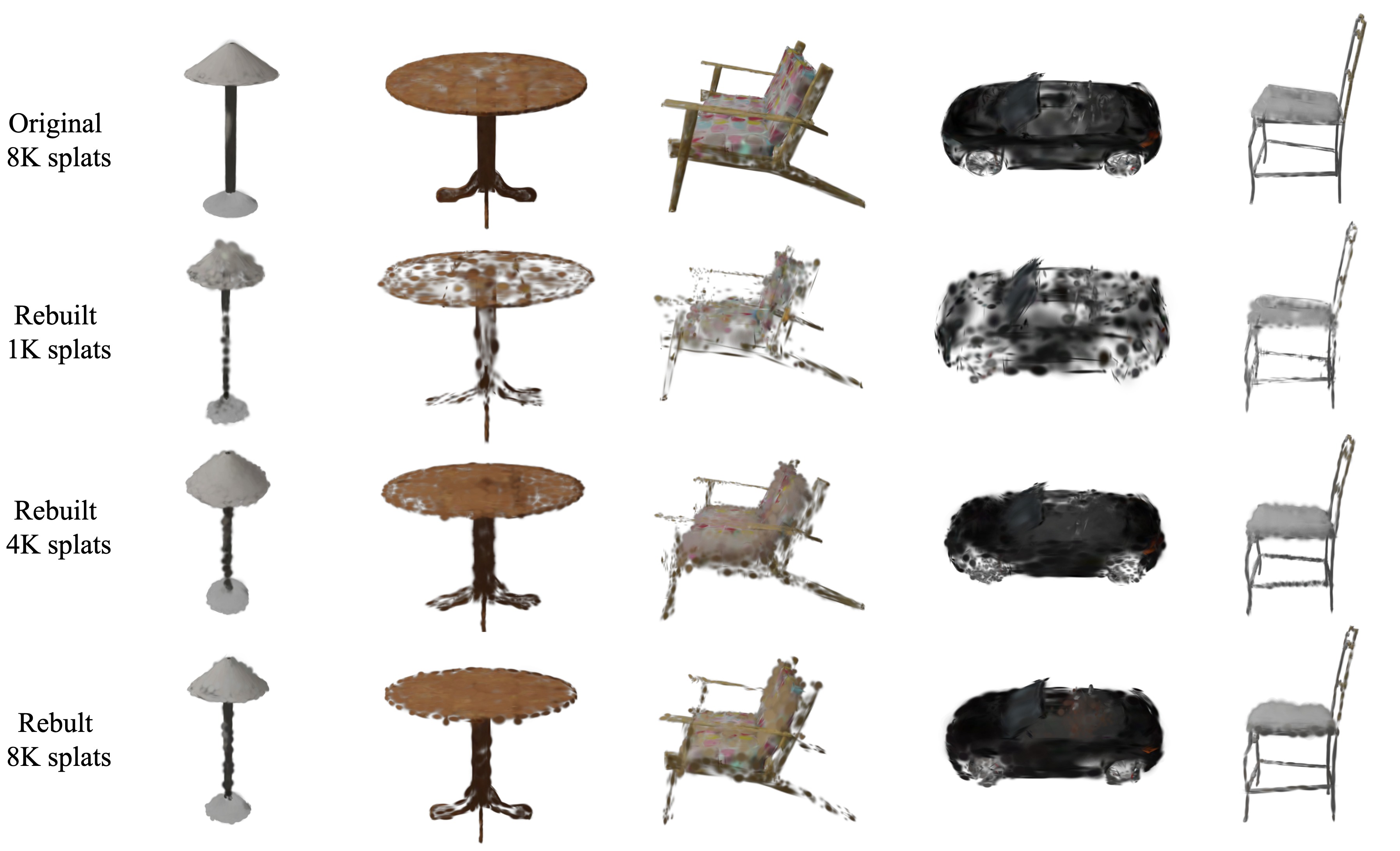}
\caption{\textbf{Qualitative Comparison of Reconstructed Gaussian Splats with Different Total Numbers}. We compare the reconstruction results with 1024, 4096, and 8192 Gaussian splats. As the number of splats increases, the reconstructions by \ours capture notably more color and geometric details, such as the colors of the bench and the frames of the chair, highlighting the importance of denser splat inputs as the finer color and geometry boost the downstream task performance.}
\label{fig:compare_recon_1k_8k}
\end{figure*}

\noindent\cref{fig:vis_recon} presents the reconstructed Gaussian splats using the \ours model with input of 4096 Gaussians during pretraining. Our model effectively rebuilds fine details, such as the foot of the table and the frames of the chair. Additionally, \cref{fig:compare_recon_1k_8k} compares the reconstructions on the same objects but with $1K$, $4K$, and $8K$ Gaussians. The ones with $8K$ Gaussians clearly capture more color and geometric details. Not only does pretraining on denser inputs result in better reconstructions, but it also boosts performance on downstream tasks (\cf~\cref{tab:ablation_pretraining}).

\section{Ablation on Pretraining Splats Number}
Throughout our experiments in the main paper, we use an input number of $1K$ for pretraining. \cref{tab:ablation_pretraining} ablates different input number of Gaussians \wrt downstream task performance. Pretraining using $8K$ Gaussians leads to the best results on ModelNet10 classification and ShapeNet-Part segmentation. Compare to training from scratch, the pretraining stage significantly boost the downstream task performance.
\input{tabs/ablation_pretraining_number}

\section{Details on ScanObjectNN Experiments}
In the main paper we report the generalization performance of our pretrained model on the real-world point clouds in ScanObjectNN~\cite{uy2019revisiting} dataset. We didn't report the performance on the Gaussian splatted objects as the segmented mesh is not provided but only point clouds in ScanObjectNN dataset, thus it's not possible to render views for the objects and train the Gaussian splats. Given \ours already outperforms the baseline~\cite{PointMAE} in object-only classification, we expect a larger performance gain when finetuned using Gaussian splats.

\section{Future Work}
While our proposed method successfully employed a self-supervised learning strategy via MAE, achieving competitive results compared to point cloud counterparts, there are several promising avenues for further exploration based on our dataset. One direction is to delve deeper into the established self-supervised learning paradigm by leveraging Gaussian attributes to enhance the encoder's informativeness for downstream tasks. Also, exploring ways to utilize all Gaussian splats without downsampling is important, as downsampling significantly reduces reconstruction quality, as shown in \cref{fig:compare_recon_1k_8k}. Additionally, transferring knowledge from 2D foundation models into the Gaussian domain presents an intriguing direction. Furthermore, unlike image or point cloud data, which have seen integration with large language models (LLMs), exploring similar trends to integrate LLMs into the 3DGS field could yield valuable insights.

\section{Impact Statement}
The introduction of the ShapeSplat dataset and the Gaussian-MAE model mark new advancements in 3D representation and understanding. ShapeSplat, a large-scale dataset with around 206K objects across 87 categories, was created using 3.8 GPU years of computation on a TITAN XP GPU, providing a robust resource for research in 3D Gaussian Splatting (3DGS). Our work facilitates unsupervised pretraining and supervised finetuning for classification and segmentation tasks, revealing critical insights into the distribution of optimized Gaussian parameters and their differing impacts on these tasks. To fully exploit the contribution of gaussian parameters space, we introduce Gaussian feature grouping and splats pooling layers, which effectively embed similar Gaussians. By making our dataset and model publicly available, we aim to drive further research in 3D representation learning, enabling the community to explore and expand upon our work.

%% file: tabs/nomenclature.tex
\nomenclcustom{G}{01}{$C$}{Center of Gaussian splats}
\nomenclcustom{G}{02}{$O$}{Opacity of Gaussian splats}
\nomenclcustom{G}{03}{$S$}{Scale of Gaussian splats}
\nomenclcustom{G}{04}{$R$}{Rotation as quaternions of Gaussian splats}
\nomenclcustom{G}{05}{$SH$}{Spherical harmonics of Gaussian splats}
\nomenclcustom{G}{06}{$X$}{Gaussian splats, $X = [C, O, S, R, SH]$}
\nomenclcustom{G}{07}{$CT$}{Center splats obtained by Furthest Point Sampling}

\nomenclcustom{S}{01}{$N$}{Splats number of the original splatted object}
\nomenclcustom{S}{02}{$p$}{Splats number after downsampling}
\nomenclcustom{S}{03}{$n$}{Groups numbers splats are splitted into}
\nomenclcustom{S}{04}{$k$}{Per-group splats number}
\nomenclcustom{S}{05}{$r$}{Mask ratio of the splats groups}

\nomenclcustom{M}{01}{$G(\cdot)$}{Grouping feature: selected Gaussian parameters for computing the distance}
\nomenclcustom{M}{02}{$E(\cdot)$}{Embedding feature: selected Gaussian parameters served as the input for masked autoencoder}
\nomenclcustom{M}{03}{$f_{G}$}{Dimension of the grouping feature}
\nomenclcustom{M}{04}{$f_{E}$}{Dimension of the embedding feature}
\nomenclcustom{M}{05}{$T$}{Group tokens obtained from the tokenizer, $T \in \mathbb{R}^{n \times D}$}
\nomenclcustom{M}{06}{$T_{v}$}{Visible tokens after masking, $T_{v} \in \mathbb{R}^{(1-r)n \times D}$}
\nomenclcustom{M}{07}{$T_{m}$}{Masked tokens, $T_{m} \in \mathbb{R}^{rn \times D}$}
\nomenclcustom{M}{08}{$E_\text{group}$}{Concatenation of embedding feature of all the $k$ splats in a group}
\nomenclcustom{M}{09}{$\hat{E}_m$}{Reconstructed embedding feature for masked regions}
\nomenclcustom{M}{10}{$E_m$}{Masked embedding feature}
\nomenclcustom{M}{11}{$z$}{Latent variable obtained from the encoder, $z = f_{\theta}(T_{v})$}
\nomenclcustom{M}{12}{$f_{\theta}(\cdot)$}{Encoder}
\nomenclcustom{M}{13}{$g_{\phi}(\cdot)$}{Decoder}
\nomenclcustom{M}{14}{$T_l$}{Learnable token that is concatenated with $z$}
\nomenclcustom{M}{15}{$\hat{T}_m$}{Recovered masked token, $\hat{T}_m=g_{\phi}(z \oplus T_l)$}
\nomenclcustom{M}{16}{$\Phi$}{Projector outputting the recovered embedding feature}
\nomenclcustom{M}{17}{$\mathcal{\tau}$}{True distribution function from which $E$ is sampled}
\nomenclcustom{M}{18}{$\mathbb{E}[\cdot]$}{Expectation operator}
\nomenclcustom{M}{19}{$\mathcal{L}_\text{recon}$}{Reconstruction loss calculated on embedding feature $E$}

%% file: tabs/model_params.tex
\begin{table}[htb]
\centering
\scriptsize
\renewcommand{\arraystretch}{1.0}
\begin{tabularx}{\columnwidth}{lcccc}
\toprule
\textbf{Model} & \textbf{Total (M)} & \textbf{Trainable (M)} & \textbf{GFLOPs} & \textbf{GMACs} \\
\midrule
Pretrain & 28.79 & 28.79 & 93.77 & 46.16 \\
Finetune (full) & 21.78 & 21.78 & 235.06 & 116.65 \\
Finetune (mlp3) & 21.78 & 0.267 & 235.06 & 116.65 \\
Finetune (linear) & 21.52 & 0.008 & 235.04 & 116.64 \\
\bottomrule
\end{tabularx}
\caption{\textbf{Model Parameters and Computation Counts.} The table reports the total and trainable parameters (in millions), as well as GFLOPs and GMACs for the pretraining and finetuning.}
\label{tab:model_params}
\end{table}

%% file: tabs/hyper_paramters.tex
\begin{table*}[t]
    \centering
    \vskip 0.10in
    \begin{tabularx}{\linewidth}{lXXXX}
     & \texttt{Pre-training} & \multicolumn{2}{c}{\texttt{Classification}} & \texttt{Segmentation}\\
     \toprule
     Config & ShapeNet~\cite{shapenet2015} & ScanObjectNN~\cite{uy2019revisiting} & ModelNet~\cite{wu20153d} & ShapeNetPart~\cite{yi2016scalable}\\
     \midrule
     optimizer & AdamW & AdamW & AdamW & AdamW\\
     learning rate & 1e-3 & 5e-4 & 5e-4 & 1e-4 \\
     weight decay & 5e-2 & 5e-2 & 5e-2 & 5e-2 \\
     learning rate scheduler & cosine & cosine & cosine & cosine \\
     training epochs & 300 & 300 & 300 & 300\\
     batch size & 128 & 32 & 128 & 128 \\
     \midrule
     number of splats & 1024 & 2048 & 1024 & 2048 \\
     number of splats groups & 64 & 128 & 64 & 128 \\
     splats group size & 32 & 32 & 32 & 32 \\
     \midrule
     augmentation & Scale\&Trans & Scale\&Trans & Scale\&Trans & 
     Scale\&Trans \\
     \midrule
     GPU device & 1 A6000 (48G) & 1 A6000 (48G) & 1 A6000 (48G) & 1 A6000 (48G) \\
    \bottomrule
    \end{tabularx}
    \caption{\textbf{Hyperparameter Recipes for Pretraining and Finetuning}.}
    \label{tab:hyper_params}
\end{table*}

%% file: tabs/gaussian_recon_error.tex
\begin{table*}[htb]
    \vspace{2em}
    \centering
    \renewcommand{\arraystretch}{1.1}
    \scalebox{0.95}{
    \begin{tabular}{llcccccc}
        \toprule
        \multirow[c]{2}{*}{Embedding Feature} & \multirow{2}{*}{Grouping Feature} & \multicolumn{5}{c}{Reconstruction Error} & \multirow[c]{2}{*}{\makecell{ModelNet10\\ Accuracy (\%)}} \\
        \cmidrule(r){3-7}
         &  & centroid & opacity & scale & rotation & sh & \\
        \midrule
        \multicolumn{8}{l}{\textit{Grouping only by Gaussian Center}}\\
        \midrule
        center & center & 2.27  & - & -  & - &  - &  93.72 \\
        center + opacity & center  & 2.53  & 0.22 & -  & - &  - &  93.83 \\
        center + sh & center  & 2.71  & - & -  & - &  0.361 &  93.83 \\
        center + scale + rotation & center  & 2.53  & - & 0.0128  & 0.126 &  - &  94.27 \\
        all & center  &  2.70  & 0.20 & 0.0156  & 0.132 &  0.364 &  95.37 \\
        \midrule
        \multicolumn{8}{l}{\textit{Grouping only by Gaussian Center, use Splats Pooling Layer}}\\
        \midrule
        
        center & center & 2.32  & - & -  & - &  - &  94.71 \\
        center + opacity & center  & 2.60  & 0.19 & -  & - &  - & 94.82  \\
        center + sh & center  & 3.11  & - & -  & - & 0.050  &  94.82 \\
        center + scale + rotation & center  & 2.82  & - & 0.0101  & 0.125 &  - &  95.37 \\
        center+opacity+scale+rotation & center & 2.63 &0.19 &0.0097 & 0.122& - & 95.82\\
        all & center  & 2.72  & 0.19 & 0.0098  &0.127  & 0.047  & 95.48  \\
        
        \midrule
        \multicolumn{8}{l}{\textit{Grouping in Gaussian Feature Space, without Center}}\\
        \midrule
        opacity & opacity  & -  & 0.007 & -   & - &  - &  51.32 \\  

        scale + rotation &  scale + rotation  & -  & - & 0.0089  & 0.047 &  - &  90.86 \\  
        sh & sh  & -  & - & -  & - &  0.063 &  78.96 \\
        opacity+scale+rotation+sh & opacity+scale+rotation+sh  & -  & 0.058 & 0.0106 & 0.069 &  0.166 & 92.29  \\ 
        \midrule
        \multicolumn{8}{l}{\textit{Grouping in Gaussian Feature Space, with Center}}\\
        \midrule
        center + opacity & center + opacity  & 5.12  & 0.05 & -  & - &  - &  94.60 \\ 
        center + scale + rotation & center + scale + rotation  &  5.76 & - & 0.010  & 0.066 &  - &   94.89 \\
        center + sh& center + sh   &  3.50 & - & -  & - & 0.118  &  95.04 \\

        all & all  &  9.19 & 0.08 & 0.011  & 0.082 &  0.201 &  95.26  \\
        
        \bottomrule
        
    \end{tabular}
    }
    \caption{\textbf{Per-attribute Reconstruction Error of Different Grouping Methods}.}
\label{tab:recon_error}
\end{table*}

%% file: tabs/ablation_pretraining_number.tex
\begin{table}[htb!]
    \centering
    \scriptsize
    \begin{tabularx}{\columnwidth}{lCCC}
    \toprule
    Input & ModelNet10 & ModelNet40 & ShapeNet-Part \\
    \midrule
    N\textbackslash A & 93.39  & 91.44  &  83.2  \\
    1024 & \nd95.37  & \fs93.35   & \nd86.0 \\
    2048 & \nd95.37  &  \nd92.95   & 85.8 \\
    4096 & 95.26  & 92.58    & \rd85.9 \\
    8192 & \fs95.70  & \rd92.70 & \fs86.1\\
 
    \bottomrule
    \end{tabularx}
    \caption{\textbf{Ablation on Number of Splats in Pretraining} (overall accuracy for ModelNet10 and ModelNet40, instance mIoU for ShapeNet-Part). We increase the total number of Gaussian splats from 1024 to 2048, 4096, and 8192 and the number of splats groups by the respective multiples during pretraining. N\textbackslash A refers to training from scratch, \ie, without pretraining stage. Evidently, pretraining stage significantly boost the downstream task performance.
    }
\label{tab:ablation_pretraining}
\end{table}